\documentclass[sigconf, nonacm]{acmart}

\usepackage{lscape}
\usepackage{array} 
\usepackage[linesnumbered,vlined,ruled,noend]{algorithm2e}
\usepackage{algorithm2e}
\usepackage{pifont}
\usepackage{placeins}
\usepackage{float}

\newcommand{\cmark}{\ding{51}} % checkmark
\newcommand{\xmark}{\ding{55}} % cross

\usepackage{wrapfig}
\begin{document}
%\title{KeyVideoLLM:  Towards Large-scale Keyframe Video Selection for Multimodal LLMs}
\title{PAS: Data-Efficient Plug-and-Play Prompt Augmentation System}
%%
%% The "author" command and its associated commands are used to define the authors and their affiliations.
% \author{Hao Liang$^{\dagger\ddagger}$, Jiapeng Li$^{\dagger\S}$, Tianyi Bai$^*$, Chong Chen$^\diamond$, Conghui He$^*$, Bin Cui$^{\ddagger}$, Wentao Zhang$^{\ddagger}$}
% % \author{Yangyu Tao$^\ddagger$, Bin Cui$^\dagger$}
% \affiliation{
% ~~~~~$^\ddagger$Peking University~~~~~$^\S$The Open University of China~~~~~$^\diamond$Huawei Cloud BU~~~~~$^*$Shanghai AI Laboratory
% % $^*$ Corresponding Author
%  }
% \affiliation{
% $^\ddagger$hao.liang@stu.pku.edu.cn, $^\ddagger$jasper\_li@alumni.pku.edu.cn, $^\ddagger$\{bin.cui, wentao.zhang\}@pku.edu.cn,
% % $^\diamond$chenchong55@huawei.com
% }

% \renewcommand{\authors}{Hao Liang, Jiapeng Li, Tianyi Bai, Chong Chen, Conghui He, Bin Cui, Wentao Zhang}

\author{Miao Zheng$^{2\dagger}$, Hao Liang$^{1\dagger}$, Fan Yang$^{2}$, Haoze Sun$^{2}$, Tianpeng Li$^{2}$, Lingchu Xiong$^{2}$, Yan Zhang$^{2}$, Youzhen Wu$^{1, 2}$, Kun Li$^{2}$, Yanjun Shen$^{2}$, Mingan Lin$^{2}$, Tao Zhang$^{2}$, Guosheng Dong$^{2}$, Yujing Qiao$^{2}$, Kun Fang$^{2}$, Weipeng Chen$^{2}$, Bin Cui$^{1}$, Wentao Zhang$^{1*}$, Zenan Zhou$^{2*}$}
%省空间，还能获得更好效果
\affiliation{
 ~~~~~$^1$Peking University~~~~~$^2$Baichuan Inc.
}
\affiliation{
\{zhengmiao, zhouzenan\}@baichuan-inc.com, hao.liang@stu.pku.edu.cn, wentao.zhang@pku.edu.cn
}

\renewcommand{\shortauthors}{Miao Zheng, Hao Liang et al.}

\begin{abstract}
In recent years, the rise of Large Language Models (LLMs) has spurred a growing demand for plug-and-play AI systems. Among the various AI techniques, prompt engineering stands out as particularly significant. However, users often face challenges in writing prompts due to the steep learning curve and significant time investment, and existing automatic prompt engineering (APE) models can be difficult to use. To address this issue, we propose PAS, an LLM-based plug-and-play APE system.
PAS utilizes LLMs trained on high-quality, automatically generated prompt complementary datasets, resulting in exceptional performance. In comprehensive benchmarks, PAS achieves state-of-the-art (SoTA) results compared to previous APE models, with an average improvement of 6.09 points. Moreover, PAS is highly efficient, achieving SoTA performance with only 9000 data points. Additionally, PAS can autonomously generate prompt augmentation data without requiring additional human labor. Its flexibility also allows it to be compatible with all existing LLMs and applicable to a wide range of tasks.
PAS excels in human evaluations, underscoring its suitability as a plug-in for users. This combination of high performance, efficiency, and flexibility makes PAS a valuable system for enhancing the usability and effectiveness of LLMs through improved prompt engineering. The code is available at \url{https://github.com/PKU-Baichuan-MLSystemLab/PAS}.
\end{abstract}






\maketitle
\begingroup
\renewcommand\thefootnote{}\footnote{\noindent
$\dagger$ The first two authors have equal contributions. \\
$*$ Corresponding Author
%\rule{\columnwidth}{0.2pt} 
}
\addtocounter{footnote}{-1}
\endgroup

\begin{figure}[ht]
  \includegraphics[width=0.47\textwidth]{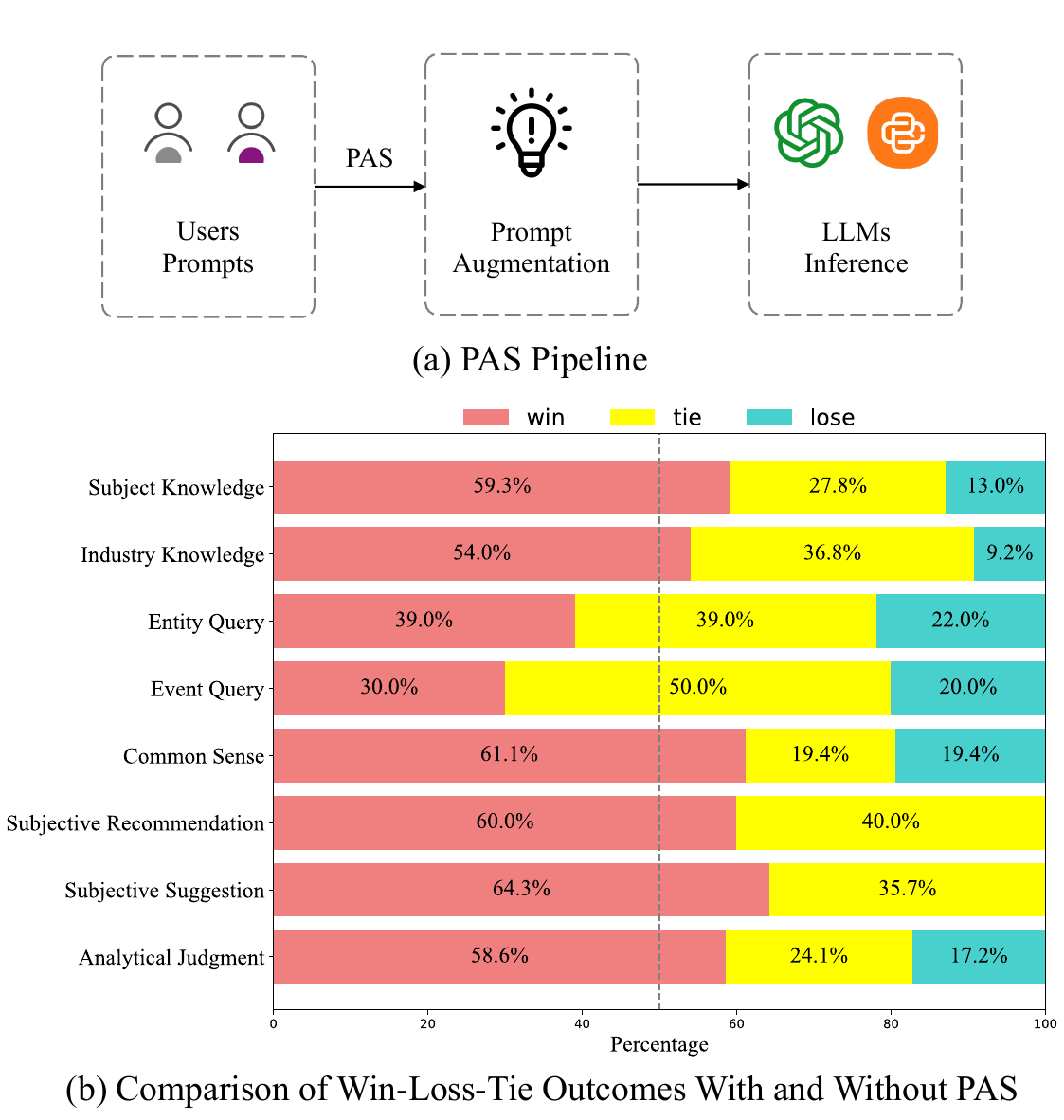}
  \caption{We first present the pipeline of the PAS in (a). PAS takes user prompts, enhances them, and then inputs the augmented prompts into LLMs. As illustrated in (b), PAS significantly improves responses across all categories in human evaluation.}
  \label{fig: Face_1}
  \vspace{-4mm}
\end{figure}
\section{Introduction}
In recent years, data management and AI systems~\cite{fernandez2023large, trummer2023bert, chen2023lingua, miao2024demystifying, nie2023flexmoe} have achieved rapid advancements and play an important role in large language models (LLMs)~\cite{chatgpt, llama}. At the same time, there is an increasing need for scalable plug-and-play systems for LLMs due to their flexibility and efficiency~\cite{al2020generalizing}.

Among the various techniques for LLMs, prompt engineering has emerged as a crucial approach due to its extremely low cost and significant enhancement of LLM performance~\cite{sahoo2024systematic}. This technique leverages the inherent capabilities of LLMs to understand and generate human-like text, enabling them to perform a wide range of applications from natural language understanding and generation to specialized domains such as medical diagnosis and legal analysis~\cite{mesko2023prompt, giray2023prompt}.
In prompt engineering techniques, few-shot learning~\cite{brown2020language} stands out because it provides a small number of examples to guide the model, thereby enhancing its task-specific performance. Chain-of-Thought (CoT)~\cite{wei2022chain} prompting guides models to reason through problems step-by-step, improving logical consistency and accuracy. In-context learning~\cite{dong2022survey}, by embedding relevant examples and instructions directly within the prompt, allows models to adapt dynamically to new tasks. 

Despite the potential of existing methods, prompt engineering is not user-friendly and requires meticulous design. Crafting effective prompts demands a deep understanding of both the model and the task at hand. This process can be time-consuming and often involves extensive trial and error to optimize performance. To tackle those weak points, automatic prompt engineering (APE) is designed for easier prompt generation~\cite{zhou2022large, pryzant2023automatic, cheng2023black}. Although APE models can automatically enhance the prompt, they have to use massive amounts of human-labeled data~\cite{cheng2023black, ouyang2022training, rafailov2024direct}. Additionally, previous methods failed to construct a flexible, user-friendly and effective APE model~\cite{zhou2022ape, pryzant2023apo}. They face the following three key challenges:

\textbf{C1. Low Efficiency.} Previous works on APE primarily use extensive human-labeled instruction datasets, which result in significant human labor~\cite{ouyang2022training, cheng2023black, rafailov2024direct}. Additionally, some methods require training a specific APE model for each LLM, leading to a substantial waste of computational resources~\cite{pryzant2023automatic, yang2023opro}.

\textbf{C2. Low Flexibility.} Previous works primarily focus on the performance of APE models, overlooking the importance of flexibility, specifically their model-agnostic and task-agnostic capabilities~\cite{pryzant2023automatic, yang2023opro}. Low-flexibility APE systems can lead to computational waste and hinder their application on LLMs~\cite{pryzant2023automatic}. Additionally, these low-flexibility systems have limited applicability across various scenarios, making them less versatile for diverse use cases and reducing their overall effectiveness in practical applications.

\textbf{C3. Poor Effectiveness.} Although several works aim to automatically enhance the quality of prompts, they often rely on outdated evaluation metrics and do not consistently yield significant improvements across various benchmarks and models~\cite{cheng2023black, ouyang2022training}. Additionally, these models fail to include human evaluations, thus lacking valuable feedback from human users~\cite{cheng2023black, pryzant2023automatic}.

\begin{figure}[ht]
  \includegraphics[width=0.47\textwidth]{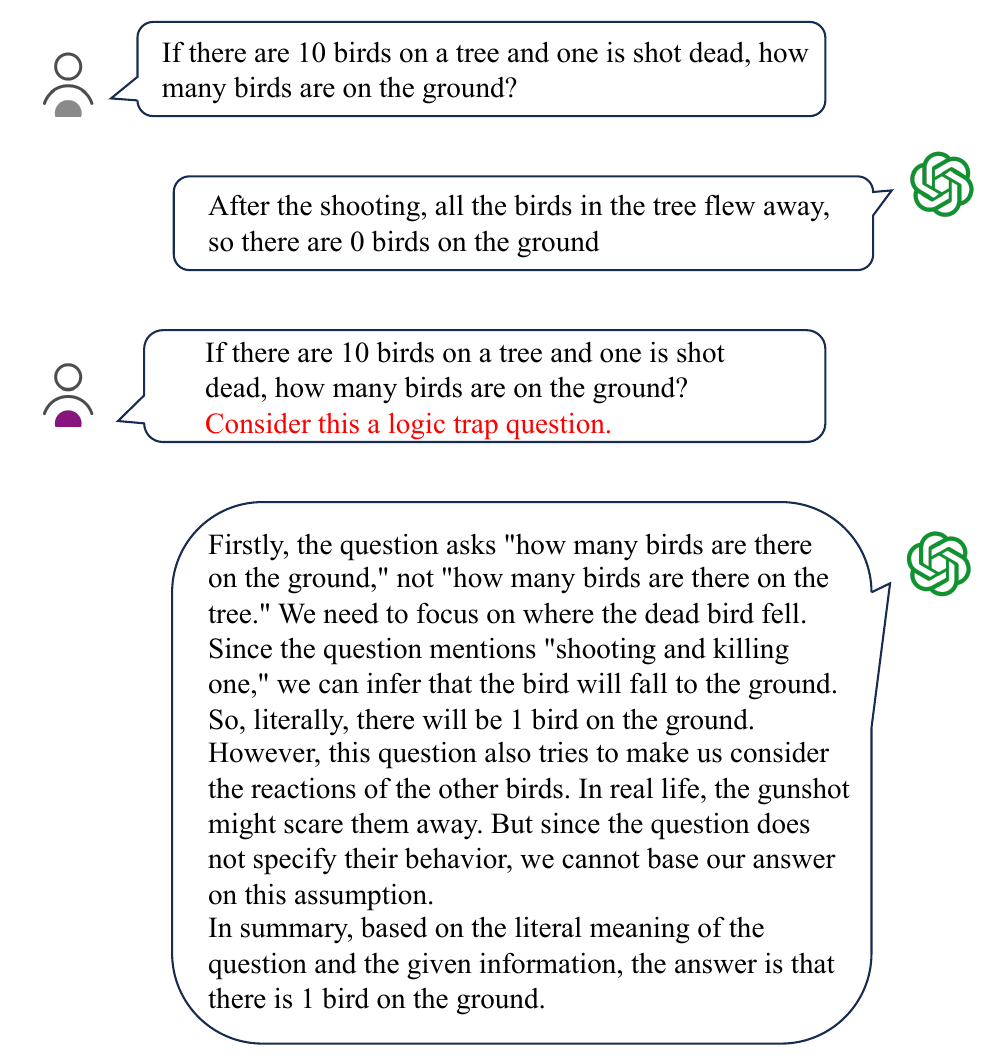}
  \caption{Case Study 1, \textcolor{red}{Red text} is the complementary prompt generated by PAS. We can see PAS can give complementary prompt to avoid logic traps.}
    \label{fig: Case_1}
    \vspace{-2mm}
\end{figure}

To address these issues, as shown in Figure \ref{fig: Face_1}(a), we propose PAS, an automatic prompt-enhancing plug-and-play system. Our approach involves two main phases: high-quality prompt selection and automatic complementary prompt generation, followed by fine-tuning of large language models (LLMs).
In the prompt selection phase, we begin by using embedding models to extract features from the prompt data. We then apply clustering algorithms to group and deduplicate similar prompts. Following this, we use LLMs to select high-quality prompts and classify them into various categories.
In the automatic complementary prompt generation phase, we employ few-shot learning techniques to generate new prompts. These generated prompts undergo a rigorous selection and regeneration process to ensure their quality. The high-quality generated data is then used to fine-tune the LLMs, forming the core of the PAS system.
%We first use embedding models to extract the feature and then use cluster algorithm to duplicate. Then we use a LLM to select high quality prompts

The core contributions of this paper are summarized as follows:
\begin{itemize}
\item \textbf{New Perspective.} To the best of our knowledge, this work is the first to construct a curated prompt complementary dataset without human labor. Additionally, we are the first to utilize this dataset to train LLMs to construct the PAS to automatically complement user prompts instead of directly modifying them.

\item \textbf{New Method.} We propose a new method that leverages diversity and quality criteria for prompt data selection. By curating a prompt dataset, we integrate few-shot learning with quality control to construct complementary prompt data. Our method, which employs this curated complementary prompt data to fine-tune LLMs, facilitates effective, efficient, and flexible prompt enhancement. 
%Then, few-shot learning is used for prompt complementary dataset construction. We employ GPT to identify and regenerate incorrectly generated data until it is correct. Instead of using human preference data, we only use the prompt complementary dataset to fine-tune LLMs and achieve SoTA performance.

\item \textbf{SoTA Performance.}
\textbf{(1) \textit{High Efficiency.}} Our PAS model requires only 9000 pairs of prompt complementary data to fine-tune LLMs, making it extremely data-efficient. Additionally, our data generation process is entirely automatic and requires no human labor. Moreover, our PAS model can be integrated into any LLM and solve all tasks, achieving SoTA performance. Therefore, we only need to train one LLM, reducing computational costs.
\textbf{(2) \textit{High Flexibility.}} Our PAS model can be plugged into any LLM, and is model and task agnostic. It achieves SoTA performance across all models and tasks, demonstrating its exceptional flexibility.
\textbf{(3) \textit{SoTA Performance on Multiple Benchmarks.}} Our PAS model achieves SoTA performance across multiple models and comprehensive benchmarks. It also outperforms the previous SoTA model, BPO, under identical experimental settings. Furthermore, it demonstrates superior performance on human evaluation metrics, as shown in Figure \ref{fig: Face_1}(b), highlighting the outstanding capabilities and potential applications of the PAS model. Figure \ref{fig: Case_1} further illustrates that our model has significant real-world application potential.
\end{itemize}

\section{Related Work}
\subsection{Automatic Prompt Engineering}
The effectiveness of large language models in various applications largely depends on the quality of the prompts used. There are already many designed prompts that can significantly enhance the performance of LLMs \cite{kojima2022large, wei2022chain, yao2024tree, besta2024graph, yang2024buffer, wang2023plan}. However, these methods that rely on manual prompt engineering are far less scalable. In the field of mathematical logical reasoning for LLMs, the Chain of Thought and its derived strategies are widely popular due to their effectiveness. Zero-shot CoT \cite{kojima2022large} is adding a simple sentence like “Let’s think step by step” at the end of questions to assist LLMs in generating reasoning steps. Instead of Zero-shot CoT, Manual-Cot \cite{wei2022chain} provides reasoning steps as few shots. Self-Consistency further improves language models’ reasoning performance by generating a diverse set of reasoning paths and choosing the most consistent answer in the final answer set. Tree of Thought (TOT) \cite{yao2024tree} and GOT \cite{besta2024graph} extend the reasoning pathway from linear to non-linear data structures by leveraging multiple LLM queries to elicit different plausible reasoning paths \cite{yang2024buffer}. Buffer of Thought (BOT) \cite{yang2024buffer} designs a series of thought-template for tasks, and for each problem, it retrieve a relevant thought-template to prompt LLMs. PS prompting \cite{wang2023plan} improves COT by encouraging LLMs to devise a plan before attempting to solve a problem. 

All the aforementioned prompting engineering strategies have been crafted by human expertise. To avoid manual effort, there is a lot of recent work to explore how to conduct automated prompt engineering \cite{zhang2022autocot, shum2023automaticcot, zhou2022ape, yang2023opro, pryzant2023apo, guo2023evoprompt, fernando2023promptbreeder}. Auto-Cot \cite{zhang2022autocot} partitions questions of a given dataset into a few clusters and generates reasoning chains to construct demonstrations for each cluster for Few-shot COT. Automatic-COT \cite{shum2023automaticcot} creates rationale chains to augment exemplars and filters out incorrect ones by checking against the ground truth. Both of them improve the performance of vanilla COT \cite{kojima2022large, wei2022chain}. Unlike previous works, OPRO~\cite{yang2023opro}, APO~\cite{pryzant2023apo}, and APE~\cite{zhou2022ape} provide an optimizer's perspective for automatically finding prompts. OPRO~\cite{yang2023opro} leverages LLMs as optimizers, using the accuracy of training datasets—unavailable in real-world scenarios—as the objective value. APO~\cite{pryzant2023apo} provides detailed guidance on prompt refinement at each step, based on the differences between responses and targets. Evoprompt~\cite{guo2023evoprompt} and Promptbreeder~\cite{fernando2023promptbreeder} introduce evolutionary algorithms (EAs) into discrete prompt optimization for specific domains. Similar to evolutionary algorithms, they require evaluating the fitness of each individual prompt in the population, presenting significant challenges in practical applications. Additionally, exploring dozens of generations of prompts imposes a considerable burden. 

% \subsection{Data Generation Methods}
% The data generation method can
\subsection{Plug-and-Play Systems}
Plug-and-play systems have garnered significant attention in recent years due to their modularity and ease of integration in various machine-learning workflows. These systems are designed to operate seamlessly with existing frameworks, allowing for quick and flexible augmentation of functionalities without the need for extensive reconfiguration~\cite{abdulrazak2006enabling, zhang2021plug, venkatakrishnan2013plug}. 

In image processing research, plug-and-play systems are commonly applied for its outstanding flexibility. Image reconstruction, denoising, deblurring, image enhancement, and super-resolution are all fields where plug-and-play systems are highly needed. By integrating various image processing modules into a unified framework, plug-and-play systems can flexibly combine different methods to achieve better image processing results. Moreover, this system allows for the easy addition or replacement of new processing modules without redesigning the entire algorithm, significantly improving the efficiency and effectiveness of image processing.

In the field of artificial intelligence, the rapid advancement of machine learning models has spurred a growing demand for plug-and-play systems. These systems enable seamless integration and adaptation of AI technologies across various applications. \citet{al2020generalizing} have underscored the critical role of AI plug-and-play systems in enhancing scalability, flexibility, and usability in modern computational frameworks. 

\subsection{Data Quality and Data Selection}
The advent of large language models has brought about a substantial increase in the volume of training data~\cite{llama, openai2023gpt}. In this scenario, the quality of data becomes paramount. LLMs, trained on vast amounts of data, can capture subtle nuances and complex patterns in language, excelling in various natural language processing tasks. However, the increase in data volume also brings new challenges, particularly in data quality and data selection~\cite{bai2024survey}. In this section, we mainly discuss the effectiveness of data quality and data selection.

\paragraph{Data Quality}: High-quality data can significantly enhance the performance of models~\cite{llama3repo}. As the volume of data increases, ensuring high data quality becomes more challenging because it requires more resources for data cleaning, selection and annotation~\cite{bai2024survey}. Poor quality data can lead to models learning incorrect patterns and making inaccurate predictions.  

\paragraph{Data Selection}: LLMs-based methods were commonly used in data selection~\cite{bai2024survey}. For instance, \citet{du2023mods} leverages DeBERTa~\cite{he2020deberta} for scoring, retaining high-quality data, and combining it with the k-center greedy algorithm to select diverse data. \citet{chen2023alpagasus} score the accuracy of data using ChatGPT to pick out high-quality data. \citet{xu2023rethinking} use GPT-4 to rewrite data to increase their complexity and then streamline it by reducing its variety and improving its quality. \citet{liu2023makes} train two models using ChatGPT's labeled data to score the quality and complexity of the data. \citet{lu2023instag} rely on ChatGPT to tag each instance, defining its complexity and diversity based on these tags. \citet{parkar2024selectllm} first cluster the data, and then use GPT-4 to select high-quality data for each cluster. 

Given the critical role of data quality and selection in enhancing model performance, our paper focuses on leveraging advanced data selection techniques to optimize prompts and their complementary data. By employing methods that integrate scoring mechanisms from LLMs and clustering techniques, we aim to efficiently identify and utilize high-quality data clusters for prompt engineering. 
%This approach not only enhances the robustness and accuracy of our model but also contributes to advancing the field's understanding of effective data-centric strategies in multimodal large language models.
\begin{figure*}[ht]
  \includegraphics[width=\textwidth]{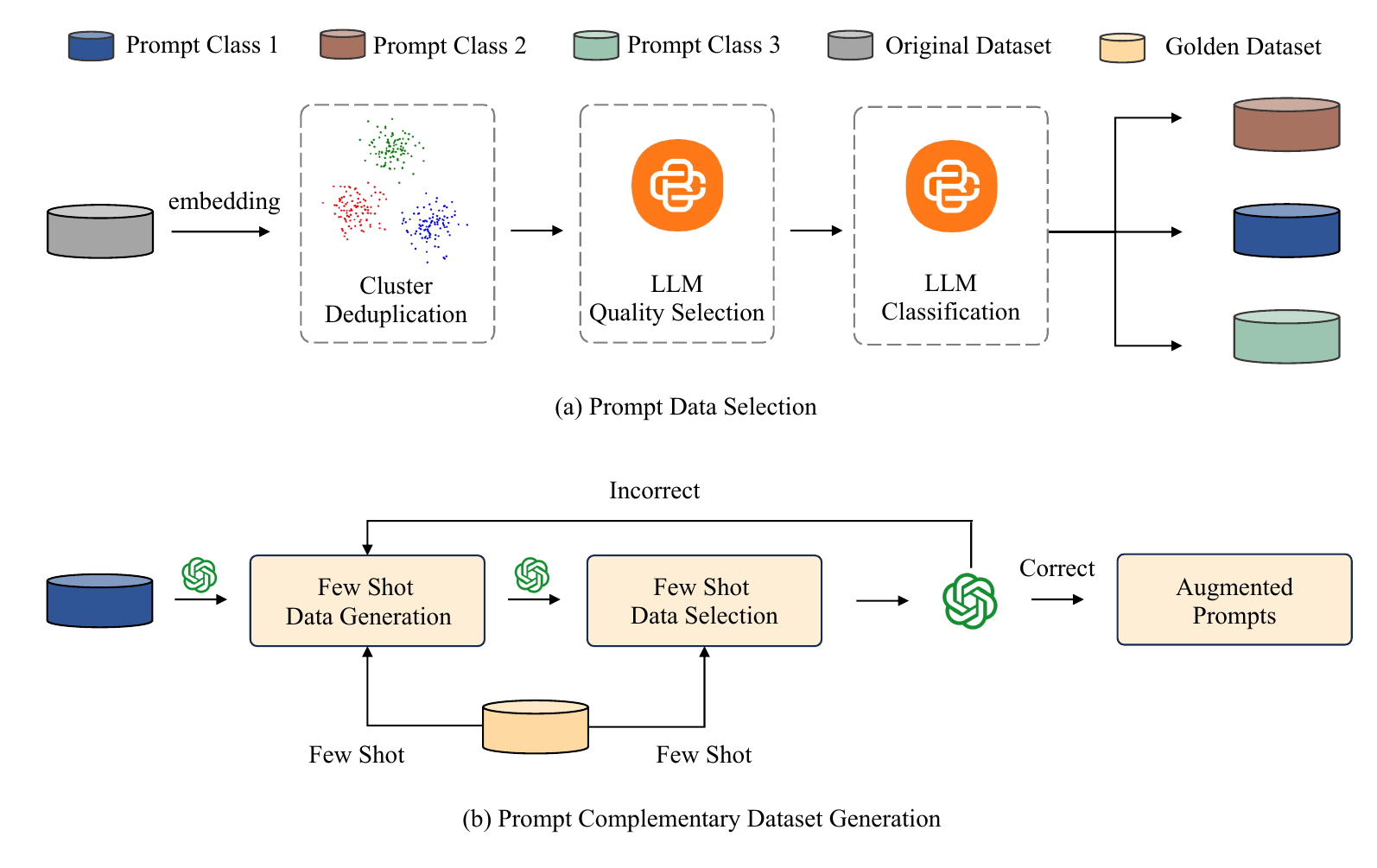}
  \caption{Pipeline for selecting prompt data and generating complementary prompt data.}
  \label{fig:Main}
\end{figure*}
\section{Method}\label{sec: Method}
In this section, we first summarize the collection and process of prompt data in section \ref{sec: Prompt Data Collection}. Then we introduce the prompt complementary data generation pipeline in section \ref{sec: Prompt Augmentation Dataset} to automatically generate high-quality prompt complementary data. After the prompt dataset is generated, we give a comprehensive analysis of the dataset in section \ref{sec: Prompt Complementary Dataset}. At last, in section \ref{sec: Prompt Plug}, we introduce utilizing the prompt augmentation dataset for LLM fine-tuning and then use the fine-tuned LLM to construct a plug-and-play APE system.

\subsection{Prompts Data Collection}\label{sec: Prompt Data Collection}
In this section, we introduce the prompt collection process. To generate high-quality (prompt, complementary prompt) pairs, we first need to select high-quality prompts. To achieve this, we use two curated datasets: the LMSYS-1M dataset~\cite{zheng2023lmsys} and the WildChat dataset~\cite{zhao2024wildchat}. We use $P$ to denote the prompt dataset. As shown in Figure \ref{fig:Main}(a), our data selection process involves three main steps to ensure the quality and relevance of the data:
\paragraph{Deduplication} Deduplication is performed to create a diverse and efficient dataset. Using the SimCSE bge embedding model, all prompts from the LMSYS-1M and WildChat datasets are embedded. 
$$P_{\text{embed}} = \text{SimCSE}(P)$$
Subsequently, the HNSW clustering algorithm is employed to group these embeddings. A small subset of data is then extracted from each cluster to reduce redundancy.

\paragraph{Quality Selection} Quality filtering is performed to select high-quality data because such data can not only reduce computational costs but also enhance the model's performance. For quality selection, the BaiChuan 13b model~\cite{yang2023baichuan} serves as the base model. We filter out low-quality entries using the formula below:
\[
Q_{\text{score}}(p_i) = \text{BaiChuan 13b}(p_i)
\]
\[
P_{\text{filtered}} = \{ p_i \in P \mid Q_{\text{score}}(p_i) \geq \tau \}
\]
Here, \( Q_{\text{score}}(p_i) \) represents the quality score assigned by the BaiChuan 13b model to prompt \( p_i \), and \( \tau \) denotes the quality threshold.
By employing quality selection, we aim to enhance the overall quality of prompt data.

\paragraph{Classification} To facilitate few-shot learning for prompt complementary dataset generation, we categorize the data into different categories. For accurate classification, we fine-tune a BaiChuan 13b model~\cite{yang2023baichuan} using 60,000 internally labeled classification data points from BaiChuan Inc. This process yields a classification model that categorizes prompts into common categories such as Q\&A and coding.

The steps of deduplication, quality selection, and classification ensure diversity, quality, and accurate categorization of the data. Approximately 9,000 high-quality classified data points are obtained through these processes. Subsequently, these 9,000 high-quality prompts are utilized to generate high-quality (prompt, complementary prompt) pairs.

%lysys-1m wildchat

\subsection{Prompts Complementary Dataset}\label{sec: Prompt Augmentation Dataset}

\begin{algorithm}
\caption{Prompt Augmentation Dataset Generation}
\label{alg:data_generation}
\SetAlgoLined
\KwIn{Golden Data $D_{\text{golden}} = \{(p_i, a_i)\}_{i=1}^N$, Prompt Set $P = \{p_j\}_{j=1}^M$}
\KwOut{Generated Data $D_{\text{generated}}$}
$D_{\text{generated}} \gets \emptyset$\;

\For{each $p_j \in P_{\text{golden}}$}{
    % 使用 FewShotLearning 生成回答
    $a_j \gets \text{FewShotGenerate}(p_j, D_{\text{golden}})$\;
    $D_{\text{generated}} \gets D_{\text{generated}} \cup \{(p_j, a_j)\}$\;
}

\For{each $(p_j, a_j) \in D_{\text{generated}}$}{
    % 检查回答是否正确
    \If{not $\text{IsCorrectPair}(p_j, a_j, D_{\text{golden}})$}{
        % 重新生成回答，直到正确
        $D_{\text{generated}} \gets (D_{\text{generated}} - \{(p_j, a_j)\})$\;
        \While{not $\text{IsCorrectPair}(p_j, a_j, D_{\text{golden}})$}{
            $a_j \gets \text{FewShotGenerate}(p_j, D_{\text{golden}})$\;
        }
        % 替换不正确的pair
        $D_{\text{generated}} \gets D_{\text{generated}} \cup \{(p_j, a_j)\}$\;
    }
}

\SetKwFunction{FewShotGenerate}{FewShotGenerate}
\SetKwFunction{IsCorrectPair}{IsCorrectPair}

\BlankLine

\textbf{Function \FewShotGenerate{$p, D_{\text{golden}}$}}: \\
\Indp
\KwIn{Prompt $p$, Golden Data $D_{\text{golden}}$} 
\KwOut{Generated Answer $a$}
% 使用 few shot learning 生成回答的具体步骤
$a \gets \text{FewShot Learning with } D_{\text{golden}}$\;
\KwRet $a$\;
\Indm
\BlankLine

\textbf{Function \IsCorrectPair{$p, a, D_{\text{golden}}$}}: \\
\Indp
\KwIn{Prompt $p$, Answer $a$, Golden Data $D_{\text{golden}}$} 
\KwOut{Boolean $isCorrect$}
% 使用 few shot learning 判断回答是否正确的具体步骤
$isCorrect \gets \text{FewShot Eval with } D_{\text{golden}}$\;
\KwRet $isCorrect$\;
\Indm

\end{algorithm}

To generate a high-quality prompt complementary dataset, we designed an automated data generation pipeline based on Few-Shot Learning. The algorithm mainly consists of two phases: data generation and data selection with regeneration. We use a set of golden data $D_{\text{golden}} = \{(p_i, a_i)\}_{i=1}^N$, containing 4 to 5 pairs of few-shot examples for each category from BaiChuan Inc. Then, we utilize the prompt dataset $P_{\text{golden}}$ from Section \ref{sec: Prompt Data Collection} to generate high-quality (prompt, complementary prompt) pairs.

\paragraph{\textbf{Data Generation}}

\begin{figure}[ht]
  \includegraphics[width=0.47\textwidth]{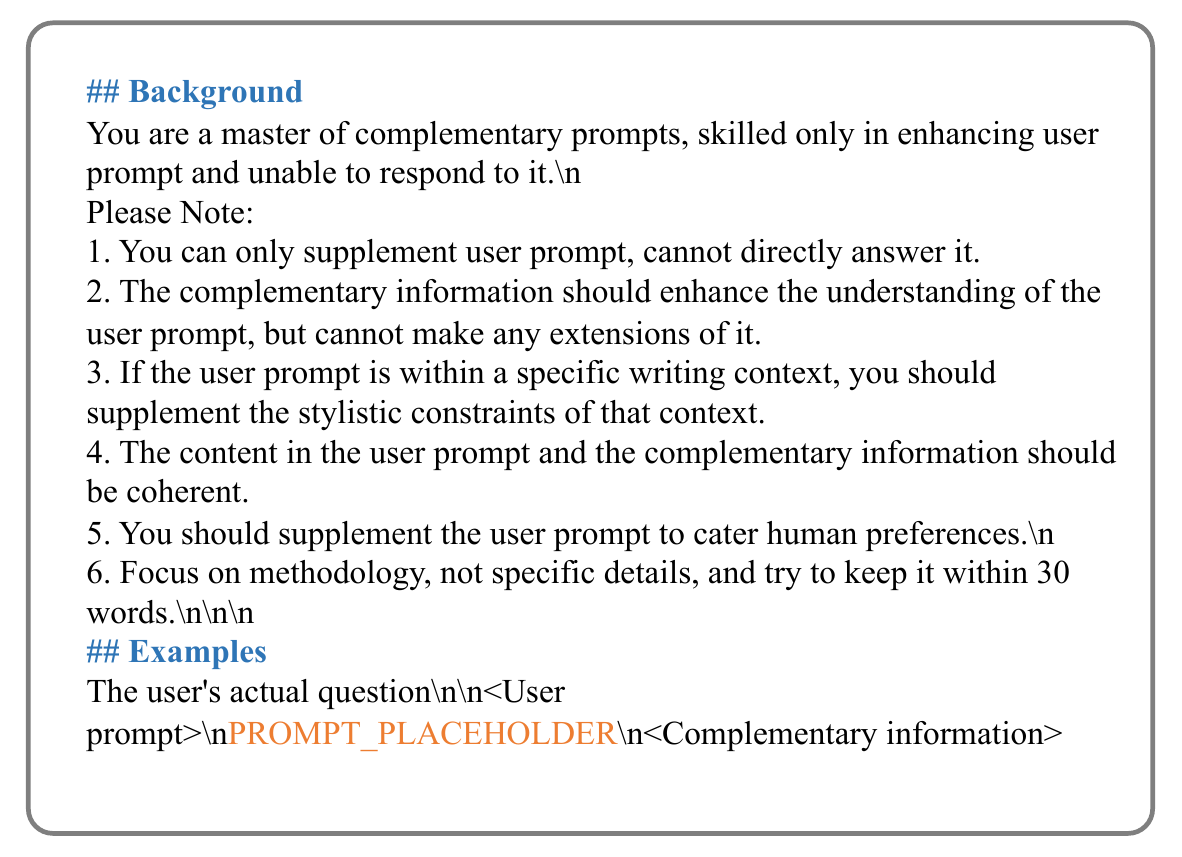}
  \caption{Complementary Dataset Generation Prompt}
  \label{fig:Prompt_1}
\end{figure}

For each prompt $p_j \in P_{\text{golden}}$ in every category, we utilize the Few-Shot Learning method based on the prompt in Figure \ref{fig:Prompt_1} to generate a corresponding complementary prompt $a_j$ based on the golden data $D_{\text{golden}}$. The generated prompt-complementary prompt pair $(p_j, a_j)$ is then added to $D_{\text{generated}}$.

\paragraph{\textbf{Data Selection and Regeneration}}
We observed that not all the generated complementary prompt data are of high quality or useful for the original prompt. To address this issue, we proposed a data selection and regeneration pipeline for high-quality complementary prompts.

\begin{figure}[ht]
  \includegraphics[width=0.47\textwidth]{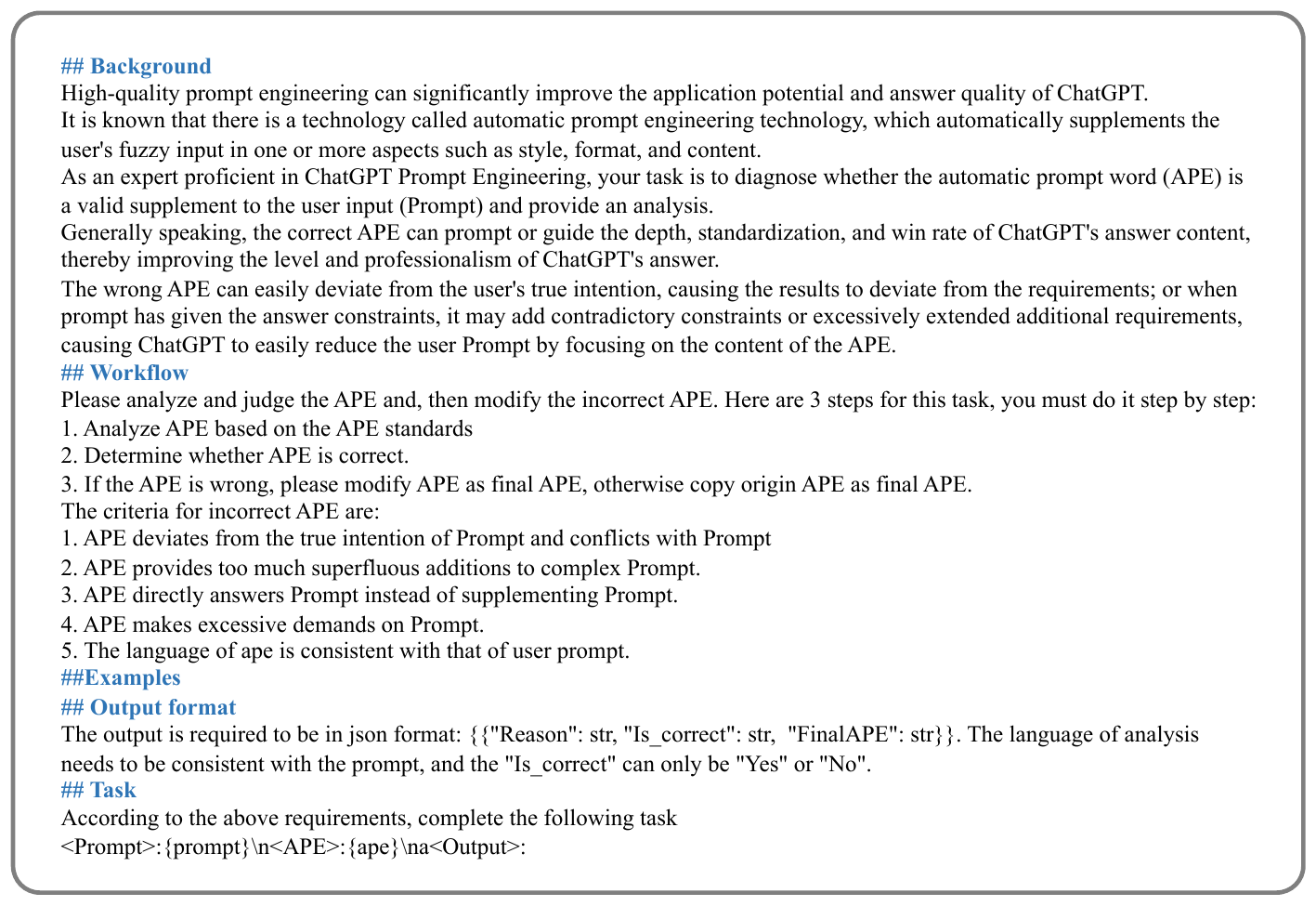}
  \caption{Data Selection and Regeneration Prompt}
  \label{fig:Prompt_2}
\end{figure}

For each generated prompt-answer pair $(p_j, a_j) \in D_{\text{generated}}$, we use Few-Shot Learning based on the prompt in Figure \ref{fig:Prompt_2} to evaluate its correctness. If the evaluation result is incorrect, we remove the pair and use Few-Shot Learning in the data generation phase to regenerate the answer until the correct answer is generated. Finally, we add the correct prompt-answer pair back to $D_{\text{generated}}$.

Through this data selection and regeneration process, we can automatically generate a prompt complementary dataset while ensuring data quality. This process provides reliable data support for subsequent model training. To better summarize the contents of this section, we translated the data selection process into an algorithm and a pipeline. The data generation pipeline is summarized in Figure \ref{fig:Main}(b) and Algorithm \ref{alg:data_generation}.

\subsection{Prompt Complementary Dataset}\label{sec: Prompt Complementary Dataset}
\begin{figure*}[ht]
  \includegraphics[width=\textwidth]{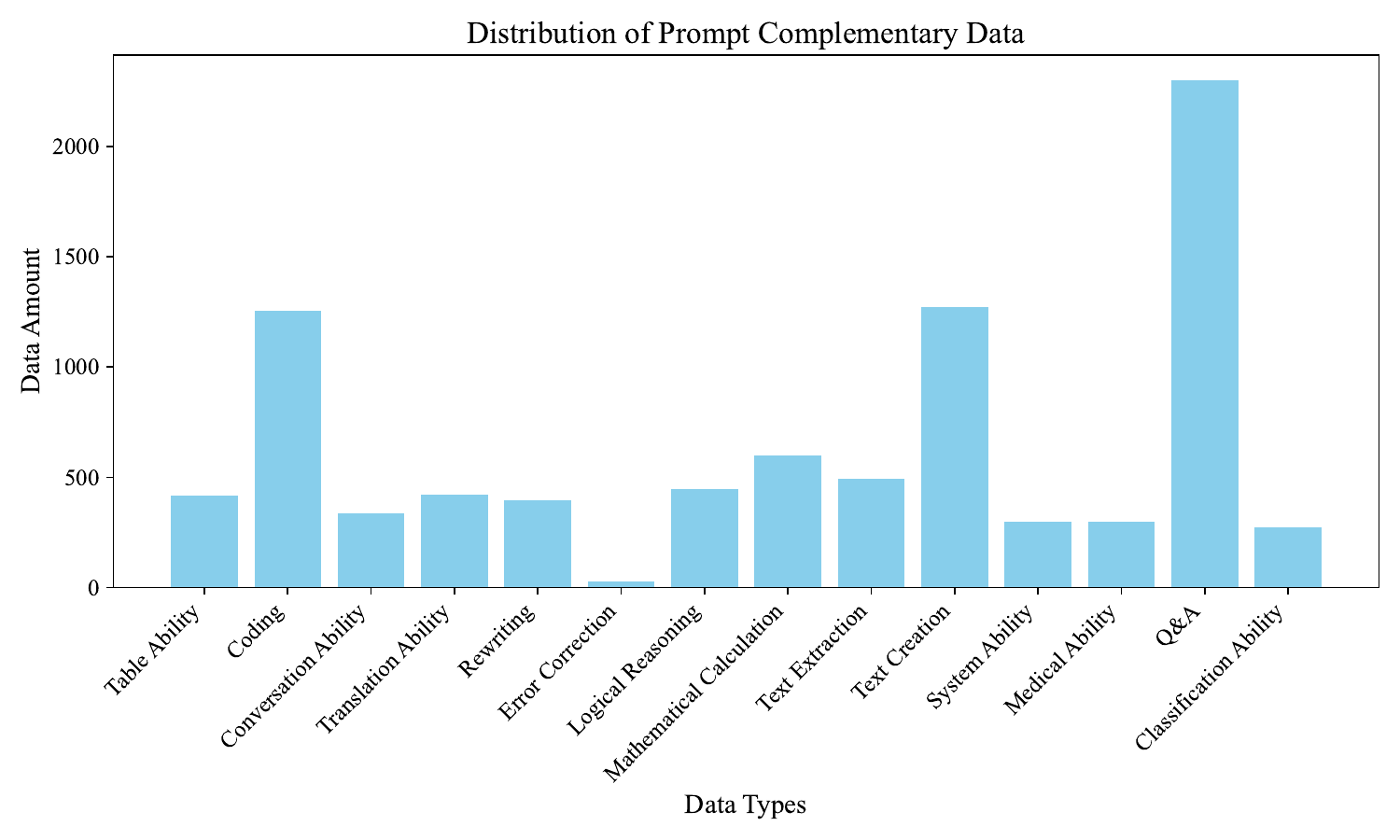}
  \caption{Prompt Complementary Dataset Distribution}
  \label{fig: Distribution}
  %\vspace{-2mm}
\end{figure*}
In this section, we present a detailed analysis of the generated prompt-complementary dataset. The dataset comprises approximately 9,000 high-quality (prompt, complementary prompt) pairs, as illustrated in Figure \ref{fig: Distribution}. Figure \ref{fig: Distribution} summarizes the distribution of the dataset, revealing that it spans 14 categories, with each category containing approximately 500 data points. This wide coverage of various prompt cases underscores the dataset's exceptional generalization capability. Moreover, there is a significant presence of Coding and Q\&A data, which are widely utilized functions, thus justifying their prominent representation.

Despite these strengths in data distribution, our automated process for generating complementary prompts, as detailed in Section \ref{sec: Prompt Augmentation Dataset}, allows us to exert control over the categories of generated data. This flexibility enables our method to cater to both general-purpose models and specialized data needs, thereby enhancing prompt adaptability across specific domains. Tailoring data generation to specific domains facilitates comprehensive training for diverse PAS tasks across various fields.

Overall, the broad coverage of various prompt cases demonstrates the dataset's exceptional generalization ability, with emphasis on critical functionalities. Additionally, our method's capability to control dataset distribution supports the development of PAS systems across all domains.

\subsection{Automatic Prompt Complementary Plug-and-Play System PAS}\label{sec: Prompt Plug}
In this section, we first use the generated prompt-complementary dataset in section \ref{sec: Prompt Augmentation Dataset} to fine-tune LLMs for prompt-complementary tasks. Given the flexibility of LLMs as automatic prompt-complementary tools that can be integrated into other generative LLMs, these prompt-complementary models can serve as an automatic, plug-and-play system to enhance LLM performance.

\paragraph{\textbf{Fine-tune LLMs for Prompt Complementary Models}}
By utilizing the high-quality generated dataset $D_{\text{generated}}$ in section \ref{sec: Prompt Augmentation Dataset}, we can automatically fine-tune LLMs to have a prompt complementary ability. We use $M_p$ to denote an automatically prompt complementary model, and $M$ to denote an LLM, which can be written as the following mathematic formula:
$$M_p \gets \text{SFT}(M; D_{\text{generated}})$$
We call this $M_e$ model PAS, a prompt complementary model with can serve as an automatic, plug-and-play system to enhance LLM performance.

\paragraph{\textbf{PAS Enhances LLMs Performance}}
For a prompt $p$ from the user, the complementary prompt $p_c$ can be generated using the following formula:
\[
p_c = M_p(p)
\]

The enhanced response $r_e$ is then given by:
\[
r_e = \text{LLMs}(\text{cat}(p, p_c))
\]

By generating a complementary prompt, the PAS can improve the user's prompt without altering the original input. As demonstrated in Section \ref{sec: Experiments}, PAS effectively enhances the performance of LLMs.

\paragraph{\textbf{Plug-and-Play LLMs Promoting System}}
PAS can be plugged into any other LLMs available via public APIs~\cite{chatgpt} and can also be integrated into models with open parameters~\cite{qwen, llama}. This flexibility allows for a wide range of applications and improvements across different platforms and systems.

The primary advantage of such a system is its ability to seamlessly enhance the capabilities of existing LLMs without the need for extensive retraining or modification. By simply augmenting the input prompts, PAS leverages the strengths of the underlying models while providing a mechanism to improve their performance. This approach is both cost-effective and efficient, enabling better utilization of computational resources and accelerating the deployment of enhanced language models in various applications.

\begin{table*}
\centering
\caption{Comparison of PAS, BPO and not using APE (baseline). We can see PAS significantly outperform the baseline with an average improvement of 8 points. Additionally, when compared to the previous state-of-the-art model, BPO, our model still exceeds it by an average of 6.09 points.}
\label{tab: Main_Exp}
\begin{tabular}{cccccccc}
\toprule
\textbf{Main Model} & \textbf{APE-model} & \textbf{Arena-hard} & \textbf{Alpaca-Eval 2.0} & \textbf{Alpaca-Eval 2.0 (LC)} & \textbf{Average} & $\uparrow$\\ 
\midrule
GPT-4-turbo-2024-04-09 & - & 76.60 & 46.12 & 55.02 & 59.25 & - \\ 
GPT-4-1106-preview & - & 74.80 & 50.00 & 50.00 & 58.27 & - \\ 
GPT-4-0613 & - & 37.9 & 15.80 & 30.20 & 27.97 & - \\ 
GPT-3.5-turbo-1106 & - & 18.90 & 9.20 & 19.30 & 15.80 & - \\ 
Qwen2-72b-Instruct & - & 48.10 & 31.70 & 39.24 & 39.68 & - \\ 
LLaMA-3-70b-Instruct & - & 41.10 & 33.18 & 34.42 & 36.23 & - \\ 
\midrule
\textbf{Average} & - & 49.57 & 31.0 & 38.03 & 39.53 & -\\ 
\midrule
GPT-4-turbo-2024-04-09 & BPO & 76.60 & 54.65 & 55.28 & 62.18 & \textcolor{red}{+2.93} \\ 
GPT-4-1106-preview & BPO & 74.60 & 55.19 & 52.91 & 60.90 & \textcolor{red}{+2.63} \\ 
GPT-4-0613 & BPO & 38.60 & 19.61 & 34.08 & 30.76 & \textcolor{red}{+2.79} \\ 
GPT-3.5-turbo-1106 & BPO & 15.90 & 10.25 & 20.29 & 15.48 & \textcolor{blue}{-0.32} \\ 
Qwen2-72b-Instruct & BPO & 44.40 & 31.25 & 39.02 & 38.22 & \textcolor{blue}{-1.46} \\ 
LLaMA-3-70b-Instruct & BPO & 45.20 & 38.92 & 39.24 & 41.12 & \textcolor{red}{+1.59} \\ 
\midrule
\textbf{Average} & BPO & 49.22 & 34.98 & 40.14 & 41.44 & \textcolor{red}{+1.91}\\ 
\midrule
GPT-4-turbo-2024-04-09 & PAS & 76.90 & 65.31 & 56.54 & 66.62 & \textcolor{red}{+7.37} \\ 
GPT-4-1106-preview & PAS & 78.80 & 65.92 & 53.63 & 66.12 & \textcolor{red}{+7.85}\\ 
GPT-4-0613 & PAS & 43.90 & 34.06 & 40.33 & 39.43 & \textcolor{red}{+11.46}\\ 
GPT-3.5-turbo-1106 & PAS & 22.10 & 15.82 & 23.31 & 20.41 & \textcolor{red}{+4.61}\\ 
Qwen2-72b-Instruct & PAS & 52.20 & 45.53 & 44.31 & 47.35 & \textcolor{red}{+7.67} \\ 
LLaMA-3-70b-Instruct & PAS & 50.30 & 45.01 & 40.52 & 45.28 & \textcolor{red}{+9.05} \\ 
\midrule
\textbf{Average} & PAS & 54.03 & 45.37 & 43.20 & 47.53 & \textcolor{red}{+8.00}\\ 
% \midrule
% GPT-4-turbo-2024-04-09 & PAS (PAS-BPO) & 76.9 & 65.31 & 56.54 & 66.62 \textcolor{red}{(+4.44)} & -\\ 
% GPT-4-1106-preview & PAS (PAS-BPO) & 78.8 & 65.92 & 53.63 & 66.12 \textcolor{red}{(+5.22)} & -\\ 
% GPT-4-0613 & PAS (PAS-BPO) & 43.9 & 34.06 & 40.33 & 39.43 \textcolor{red}{(+8.67)} & -\\ 
% GPT-3.5-turbo-1106 & PAS (PAS-BPO) & 22.1 & 15.82 & 23.31 & 20.41 \textcolor{red}{(+4.93)} & -\\ 
% Qwen2-72B-instruct & PAS (PAS-BPO) & 52.2 & 45.53 & 44.31 & 47.35 \textcolor{red}{(+9.13)} & -\\ 
% LLaMA-3-70b-instruct & PAS (PAS-BPO) & 50.3 & 45.01 & 40.52 & 45.28 \textcolor{red}{(+4.16)} & -\\ 
% \midrule
% \textbf{Average} & PAS (PAS-BPO) & 54.03 & 45.37 & 43.20 & 47.53 \textcolor{red}{(+6.09)} & -\\ 
\bottomrule
\end{tabular}
\vspace{-2mm}
\end{table*}
\section{Experiments}\label{sec: Experiments}
In this section, we first introduce the experimental setups. We then aim to answer the following questions to verify the effectiveness, efficiency, and robustness of our proposed PAS: \textbf{Q1}: Can our PAS achieve SoTA performance compared to previous SoTA methods? \textbf{Q2}: Can our PAS outperform the previous SoTA model with the same base model? \textbf{Q3}: How efficient and flexible is our model compared to previous APE models? \textbf{Q4}: Can PAS achieve SoTA performance in human evaluation, making it a user friendly system? \textbf{Q5}: Can we visualize the advantages of our method? \textbf{Q6}: Do we need data quality selection and regenerate module in our data generation pipeline?
\subsection{Experiments Setting}

\paragraph{\textbf{Datasets.}} We followed the steps in section \ref{sec: Method} and generated a dataset of 9000 high-quality pairs (prompt, complementary prompt).

\paragraph{\textbf{Models.}} For PAS models, we select several smaller models to efficiently train a PAS model. We select Qwen2-7b-Instruct~\cite{qwen}, LLaMA-2-7b-Instruct~\cite{llama}. Then we utilize our trained PAS models to some massive SoTA models, i.e. GPT-4-turbo-2024-04-09, GPT-4-1106-preview, GPT-4-0613, GPT-3.5-turbo-1106~\cite{chatgpt}, Qwen2-72b-Instruct~\cite{qwen}, and LLaMA-3-70b-Instruct~\cite{llama,llama3repo}.

\paragraph{\textbf{Baselines.}}
We compare the performance of PAS with models without PAS. Additionally, we compare the performance of PAS with the previous SoTA automatic prompt engineering method BPO~\cite{cheng2023black} to demonstrate the effectiveness of PAS. 

\paragraph{\textbf{Evaluation.}}

To evaluate the effectiveness of our PAS model, we used three comprehensive benchmarks to thoroughly assess the model's performance:

\begin{itemize}
    \item \textbf{Arena-hard}: This benchmark is designed to test the robustness of language models in handling complex and challenging scenarios. It includes tasks that require advanced reasoning, problem-solving, and understanding of nuanced language constructs. Models are evaluated based on their ability to navigate these hard scenarios and provide accurate, coherent responses. 

    \item \textbf{Alpaca-Eval 2.0}: This benchmark assesses the general performance of language models across a wide range of standard tasks. It includes a variety of question types and subject areas, ensuring a comprehensive evaluation of the model's capabilities. The Alpaca-Eval 2.0 is a standard for measuring the overall effectiveness and versatility of language models.

    \item \textbf{Alpaca-Eval 2.0 (LC)}: Alpaca-Eval 2.0 LC is a length-controlled version of AlpacaEval designed to mitigate biases related to response length in language model evaluations. By implementing length control, it reduces sensitivity to response length variations, enhancing robustness and interpretability of results. This improvement increases AlpacaEval's correlation with human judgments, as shown by its higher correlation with Chatbot Arena evaluations.
\end{itemize}

\paragraph{\textbf{Settings.}}
For Qwen2-7B-Instruct~\cite{qwen}, LLaMA-2-7B-Instruct~\cite{llama}, Qwen2-72B-Instruct~\cite{qwen}, LLaMA-2-7B-Instruct~\cite{llama}, and LLaMA-3-70B-Instruct~\cite{llama,llama3repo}, we primarily use the hyperparameters from the official repositories. For the GPT model series, we access the models via API. All experiments are conducted on a machine equipped with 8 NVIDIA H100 GPUs, a 120-core CPU, and 960GB of memory.

\subsection{Main Experiments}\label{sec: Main_Experiment}
To address \textbf{Q1}, we used Qwen2-7B-Instruct as the base model due to its outstanding performance. We subsequently used the prompt complementary data to train a PAS model and compared it to both the baseline model without the APE model and the previous state-of-the-art (SoTA) APE model, BPO~\cite{cheng2023black}. We integrated our model into multiple commonly used LLMs, including GPT-4-turbo-2024-04-09, GPT-4-1106-preview, GPT-4-0613, GPT-3.5-turbo-1106~\cite{chatgpt}, Qwen2-72b-Instruct~\cite{qwen}, and LLaMA-3-70b-Instruct~\cite{llama,llama3repo}.

The results in Table \ref{tab: Main_Exp} clearly illustrate the effectiveness of our PAS method across different models. Compared to the baseline without using APE, PAS shows significant improvements in all metrics, resulting in an average improvement of 8 points, demonstrating the benefits of incorporating prompt complementary data. For instance, in the case of GPT-4-0613, PAS improves the average score by 11.46 points compared to the baseline, highlighting its substantial impact.

Moreover, when compared to the previous state-of-the-art model BPO, our model significantly outperforms it, resulting in an average improvement of 6.09 points. Each model achieves more than a 4-point average improvement across all six base models compared to BPO, with a notable increase of 9.13 points for Qwen2-72b-Instruct, indicating a substantial improvement.

BPO is unstable and performs worse than the baseline in some cases, such as GPT-3.5-turbo-1106 and Qwen2-72b-Instruct, indicating that the previous SoTA model cannot consistently outperform the baseline. Considering our model exceeds the baseline by 8.00 points and BPO by 6.09 points, it further demonstrates the effectiveness and robustness of our PAS model.

Overall, our PAS method not only outperforms the baseline but also consistently surpasses the previous SoTA model BPO, establishing its robustness and effectiveness as a fine-tuning strategy for enhancing prompt-based learning systems. This consistent performance across various LLMs underscores the robustness of PAS and its potential to set new benchmarks in the field.

\begin{table*}[!ht]
\centering
\caption{Comparison of PAS and BPO using the same base model, LLaMA-2-7b-Instruct. The results demonstrate that PAS outperforms the BPO model consistently across all LLMs when using the same base model, LLaMA-2-7b-Instruct.}
\label{tab: Main_Exp_2}
\begin{tabular}{cccccccc}
\toprule
\textbf{Main Model} & \textbf{Method} & \textbf{Arena-hard} & \textbf{Alpaca-Eval 2.0} & \textbf{Alpaca-Eval 2.0 (LC)} & \textbf{Average} & $\uparrow$ \\ 
\midrule
GPT-4-turbo-2024-04-09 & BPO & 76.60 & 54.65 & 55.28 & 62.18 & - \\ 
GPT-4-1106-preview & BPO & 74.60 & 55.19 & 52.91 & 60.90 & - \\ 
GPT-4-0613 & BPO & 38.60 & 19.61 & 34.08 & 30.76 & - \\ 
GPT-3.5-turbo-1106 & BPO & 15.90 & 10.25 & 20.29 & 15.48 & - \\ 
Qwen2-72b-Instruct & BPO & 44.40 & 31.25 & 39.02 & 38.22 & - \\ 
LLaMA-3-70b-Instruct & BPO & 45.20 & 38.92 & 39.24 & 41.12 & - \\ 
\midrule
\textbf{Average} & BPO & 49.22 & 34.98 & 40.14 & 41.44 & - \\ 
\midrule
GPT-4-turbo-2024-04-09 & PAS & 73.54 & 62.58 & 54.03 & 63.38 & \textcolor{red}{+1.20} \\ 
GPT-4-1106-preview & PAS & 75.52 & 64.06 & 53.07 & 64.22 & \textcolor{red}{+3.32} \\ 
GPT-4-0613 & PAS & 40.13 & 33.11 & 36.70 & 36.65 & \textcolor{red}{+5.89} \\ 
GPT-3.5-turbo-1106 & PAS & 18.02 & 16.18 & 23.67 & 19.29 & \textcolor{red}{+3.81} \\ 
Qwen2-72b-Instruct & PAS & 47.91 & 40.59 & 39.99 & 42.83 & \textcolor{red}{+4.61} \\ 
LLaMA-3-70b-Instruct & PAS & 46.30 & 43.17 & 38.77 & 42.74 & \textcolor{red}{+1.62} \\ 
\midrule
\textbf{Average} & PAS & 50.24 & 43.28 & 41.04 & 44.85 & \textcolor{red}{+3.41} \\ 
\bottomrule
\end{tabular}
\end{table*}
\subsection{Effectiveness of PAS}
To address \textbf{Q2}, we fix the base model and compare our PAS method with the previous BPO~\cite{cheng2023black}. We use LLaMA-2-7b-Instruct, the same base model as BPO, and utilize the generated complementary prompt data to fine-tune LLaMA-2-7b-Instruct. We compare our model performance with BPO.

The results in Table \ref{tab: Main_Exp_2} clearly demonstrate the effectiveness of our PAS method across different models. Notably, PAS exhibits a marked improvement in performance metrics compared to BPO, exceeding the baseline by 3.41 points on average. This is particularly evident in models like GPT-4-0613, where the average score improvement is as high as 5.89 points. Even in cases where the improvement is smaller, such as Llama3-70b-Instruct, PAS still manages to outperform BPO, indicating its robustness and consistency. 

Overall, our PAS method consistently improves model performance across various evaluation models and settings, establishing its effectiveness and robustness as a fine-tuning strategy for enhancing prompt-based learning systems.
\subsection{Efficiency and Flexibility of PAS}

To address \textbf{Q3}, we compare the data efficiency and flexibility of the PAS approach with SoTA APE models. This comparison underscores PAS's effectiveness in terms of both data utilization and adaptability.

\subsubsection{Data Usage Comparison}
\begin{figure}
  \includegraphics[width=0.47\textwidth]{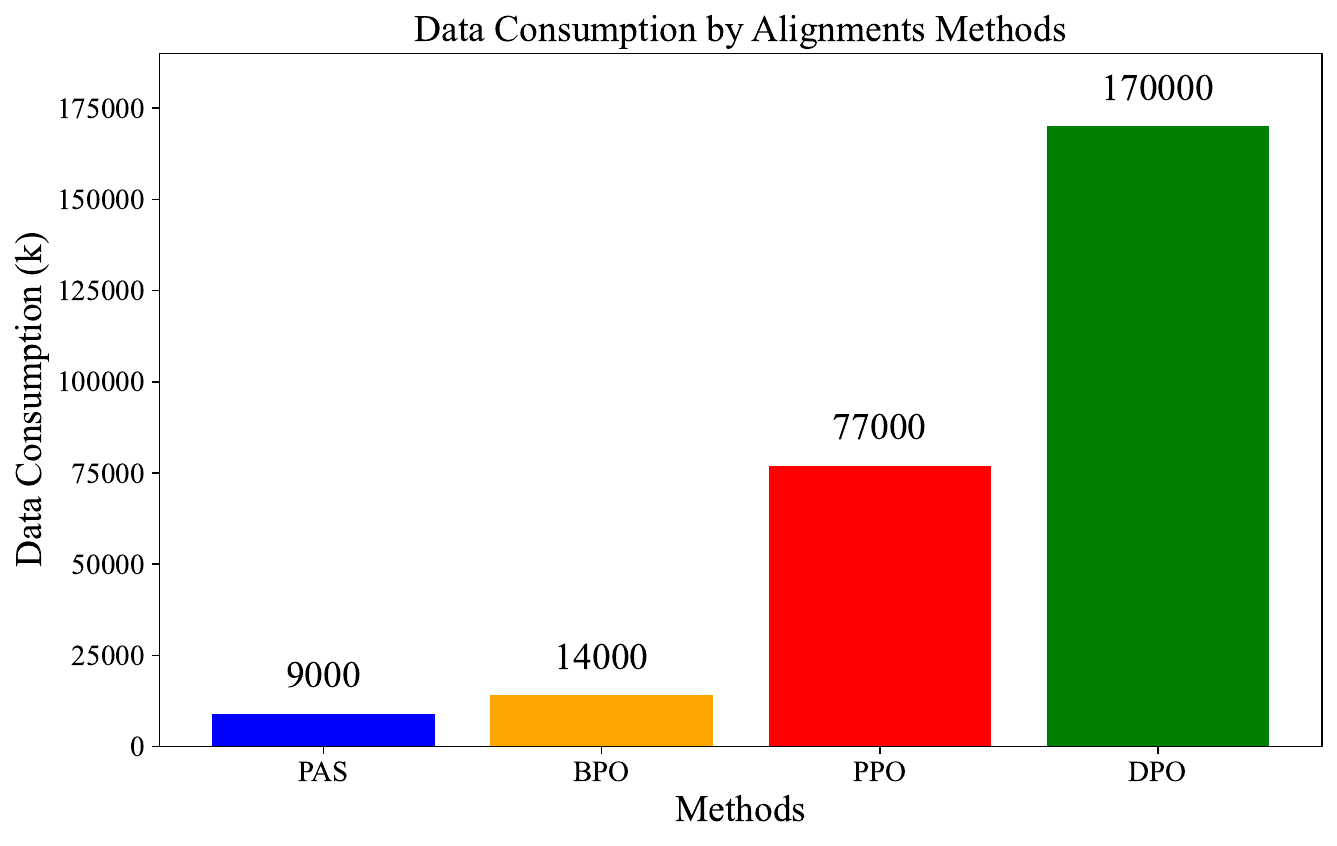}
  \caption{Efficiency of PAS Compared to Other Methods}
  \label{fig: Data_Compare}
  \vspace{-4mm}
\end{figure}

Figure~\ref{fig: Data_Compare} shows the data consumption of PAS compared to other alignment methods, including PPO, DPO, and BPO. The data consumption for each method is as follows:
\begin{itemize}
  \item PAS: 9,000
  \item BPO: 14,000
  \item PPO: 77,000
  \item DPO: 170,000
\end{itemize}

As shown in figure \ref{fig: Data_Compare}, PAS significantly reduces data consumption compared to other methods, demonstrating its efficiency in data usage. It is important to note that ProTeGi and OPRO are not task-agnostic, so they are not included in the data efficiency comparison.

We then calculated the efficiency of PAS using the formula:

\[
\text{Efficiency} = \frac{\text{Consumption}_{\text{Methods}}}{\text{Consumption}_{\text{PAS}}}
\]

Based on this formula, PAS achieves an efficiency that is 1.56 times higher than BPO, 8.56 times higher than PPO, and 18.89 times higher than DPO, highlighting its superior data efficiency.

\subsubsection{Flexibility Comparison}
Table~\ref{tab: performance_comparison} compares the need for human labor and the flexibility of PAS with other models, specifically in terms of being LLM-agnostic and task-agnostic. The table shows that PAS is both LLM-agnostic and task-agnostic, demonstrating its superior flexibility.

\begin{table}[h!]
\centering
\small
\caption{Comparison of the Need for Human Labor and Flexibility of PAS as a Plug-and-Play System}
\begin{tabular}{cccc}
\toprule
\textbf{Method} & \textbf{No Human Labor} & \textbf{LLM-Agnostic} & \textbf{Task-Agnostic} \\
\midrule
APE~\cite{guo2023evoprompt} & \xmark & \xmark & \xmark \\
Auto-Cot~\cite{zhang2022autocot} & \xmark & \xmark & \cmark \\
OPRO~\cite{yang2023opro} & \xmark & \xmark & \xmark \\
ProTeGi~\cite{pryzant2023automatic} & \xmark & \xmark & \xmark \\
BPO~\cite{cheng2023black} & \xmark & \cmark & \cmark \\
PAS & \cmark & \cmark & \cmark \\
\bottomrule
\end{tabular}
\label{tab: performance_comparison}
\vspace{-2mm}
\end{table}

\subsubsection{Summary}

The results indicate that PAS is the only method satisfying all three criteria: no need for human labor, LLM-agnostic, and task-agnostic. This underscores PAS's superior flexibility and efficiency as a plug-and-play system. Unlike other methods that require significant human intervention and have limitations in applicability across different LLMs and tasks, PAS provides a highly versatile and efficient solution.

In contrast, methods like PPO, DPO, OPRO, and ProTeGi require human labor and lack LLM-agnostic capabilities. Although BPO is both LLM-agnostic and task-agnostic, it still requires human labor, necessitating large human-annotated datasets, whereas PAS can scale up automatically. This highlights PAS's significant advancements in providing an efficient, flexible, and human-labor-free prompt augmentation system.

\subsection{Human Evaluation}
To address \textbf{Q4}, we conducted a comprehensive evaluation using human evaluators to assess the online performance of our PAS compared to the baseline model without any prompt augmentation. We compared the good same bad (GSB), availability proportion, full mark proportion, and average score.

From Figure \ref{fig: Face_1}, we can see that PAS outperforms the baseline model in various scenarios. Specifically, the results show a higher percentage of wins across different categories such as Analytical Judgment, Subjective Suggestion, Subjective Recommendation, Common Sense, Event Query, Entity Query, Industry Knowledge, and Subject Knowledge. For example, PAS achieves 58.6\% wins in Analytical Judgment, 64.3\% in Subjective Suggestion, and 61.1\% in Common Sense, demonstrating its effectiveness in enhancing performance compared to the baseline without PAS.

As shown in Table \ref{tab:human_evaluation},  our PAS consistently outperforms the baseline, achieving significant improvements in the full mark ratio, average score, and availability ratio. These enhancements across all three evaluation metrics in every benchmark demonstrate the effectiveness of our model. The results indicate not only strong performance on evaluation benchmarks but also positive feedback from human evaluators, showcasing a user-friendly model.

Furthermore, the consistent performance enhancements across all benchmarks highlight the generalization ability and robustness of our model, suggesting its applicability in various domains and its broad impact.

\begin{table*}[h!]
    \centering
    \small
    \caption{Performance Comparison of PAS vs. Non-PAS on Human Evaluation Benchmarks. The PAS consistently outperforms the non-PAS approach across various metrics.}
    \label{tab:human_evaluation}
    \begin{tabular}{l*{6}{>{\centering\arraybackslash}p{2cm}}}
        \toprule
        Benchmarks & Full Mark Proportion & Average Score & Availability Proportion & Full Mark Proportion (PAS) & Average Score (PAS) & Availability Proportion (PAS) \\
        \midrule
        Analysis and Judgment & 24.14\% & 3.84 & 91.38\% & 43.10\% \textcolor{red}{(+18.96)} & 4.21 \textcolor{red}{(+0.37)} & 94.83\% \textcolor{red}{(+3.45)} \\
        Subjective Advice & 35.71\% & 3.71 & 85.71\% & 42.86\% \textcolor{red}{(+7.15)} & 3.93 \textcolor{red}{(+0.22)} & 85.71\% \textcolor{red}{(+0.00)} \\
        Subjective Recommendation & 0.00\% & 2.4 & 60.00\% & 0.00\% \textcolor{red}{(+0.00)} & 2.8 \textcolor{red}{(+0.40)} & 80.00\% \textcolor{red}{(+20.00)} \\
        Common Sense & 5.56\% & 3.25 & 77.78\% & 27.78\% \textcolor{red}{(+22.22)} & 3.72 \textcolor{red}{(+0.47)} & 80.56\% \textcolor{red}{(+2.78)} \\
        Event Query & 20.00\% & 3.3 & 60.00\% & 30.00\% \textcolor{red}{(+10.00)} & 3.6 \textcolor{red}{(+0.30)} & 70.00\% \textcolor{red}{(+10.00)} \\
        Entity Query & 7.32\% & 3.15 & 68.29\% & 9.76\% \textcolor{red}{(+2.44)} & 3.34 \textcolor{red}{(+0.19)} & 75.61\% \textcolor{red}{(+7.32)} \\
        Industry Knowledge & 20.69\% & 3.49 & 78.16\% & 40.23\% \textcolor{red}{(+19.54)} & 3.78 \textcolor{red}{(+0.29)} & 79.31\% \textcolor{red}{(+1.15)} \\
        Academic Knowledge & 18.52\% & 3.35 & 77.78\% & 29.63\% \textcolor{red}{(+11.11)} & 3.76 \textcolor{red}{(+0.41)} & 83.33\% \textcolor{red}{(+5.55)} \\
        \midrule
        \textbf{Average} & 16.49\% & 3.35 & 77.78\% & 27.92\% \textcolor{red}{(+11.43)} & 3.76 \textcolor{red}{(+0.41)} & 83.33\% \textcolor{red}{(+5.55)} \\
        \bottomrule
    \end{tabular}
    %\vspace{-2mm}
\end{table*}

\subsection{Case Study}
To address \textbf{Q5}, this section discusses three different types of case studies that demonstrate the effectiveness of the PAS.
\paragraph{\textbf{Case Study 1}}
We examine the effectiveness of PAS in the case study presented in Figure \ref{fig: Case_1}. The query, \textit{"If there are 10 birds on a tree and one is shot dead, how many birds are on the ground?"}, is a logic trap that can easily mislead both humans and large models if no hints are given. Initially, without the assistance of PAS, GPT responded with an incorrect answer.

Our PAS approach significantly improves this prompt by providing a complementary hint to remind the model that there is a logic trap. With the assistance of PAS, the new response avoids the trap and showcases a multi-step logical reasoning process. This refined answer is clear, accurate, and guides the user through the reasoning process, explaining why only one bird would be on the ground.

From Case Study 1, we can see that PAS is capable of assisting LLMs in avoiding logic traps.
\begin{figure*}
  \includegraphics[width=\textwidth]{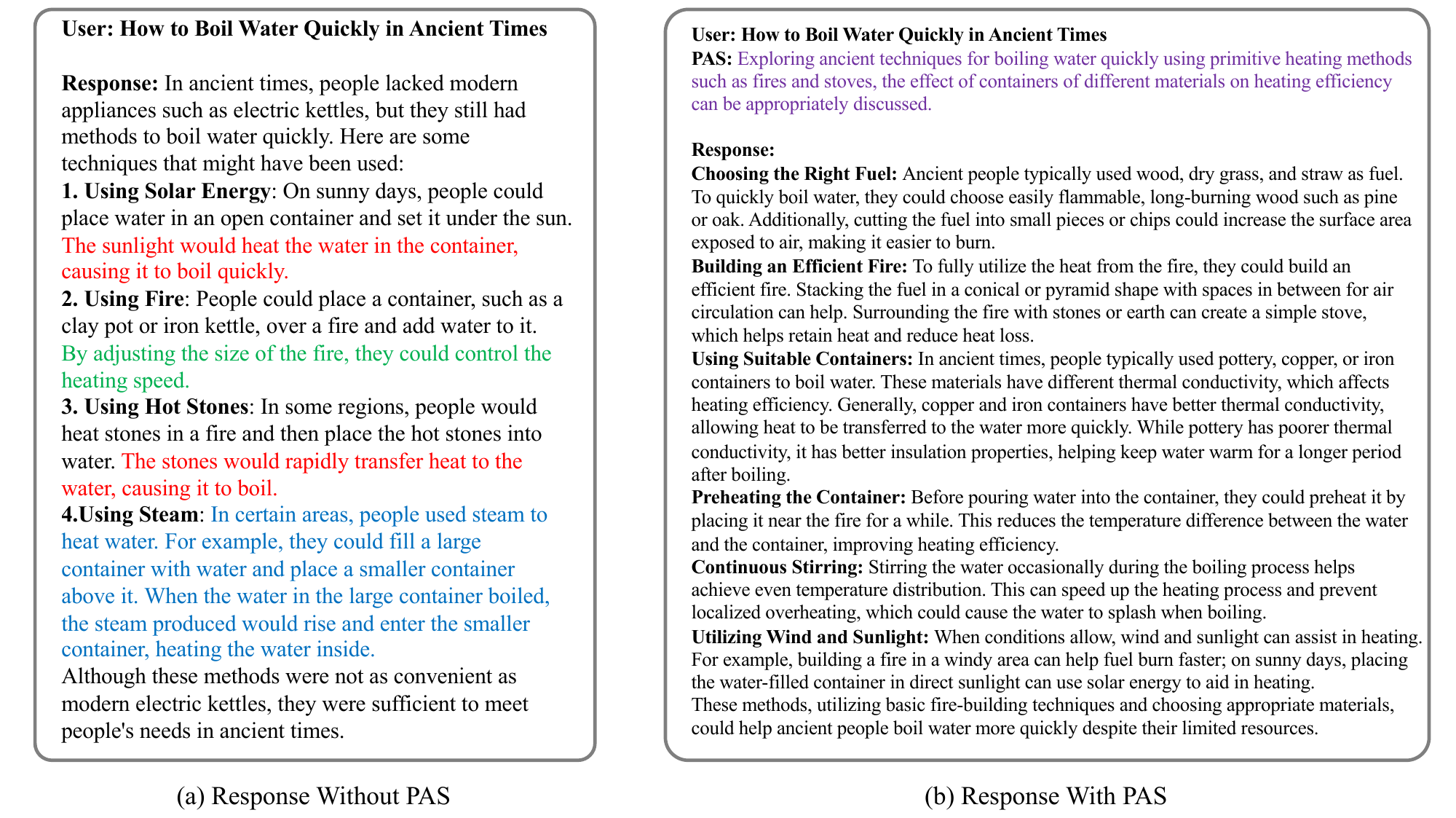}
  \caption{Case Study 2: \textcolor{red}{Red text} indicates the response is incorrect. \textcolor{green}{Green text} stands for inexplicit sentences. \textcolor{blue}{Blue text} denotes the no sense sentences. \textcolor{violet}{Purple text} is the complementary prompt generated bu PAS.}
  \label{fig: Case_2}
  \vspace{-2mm}
\end{figure*}

\begin{figure*}[!ht]
  \includegraphics[width=\textwidth]{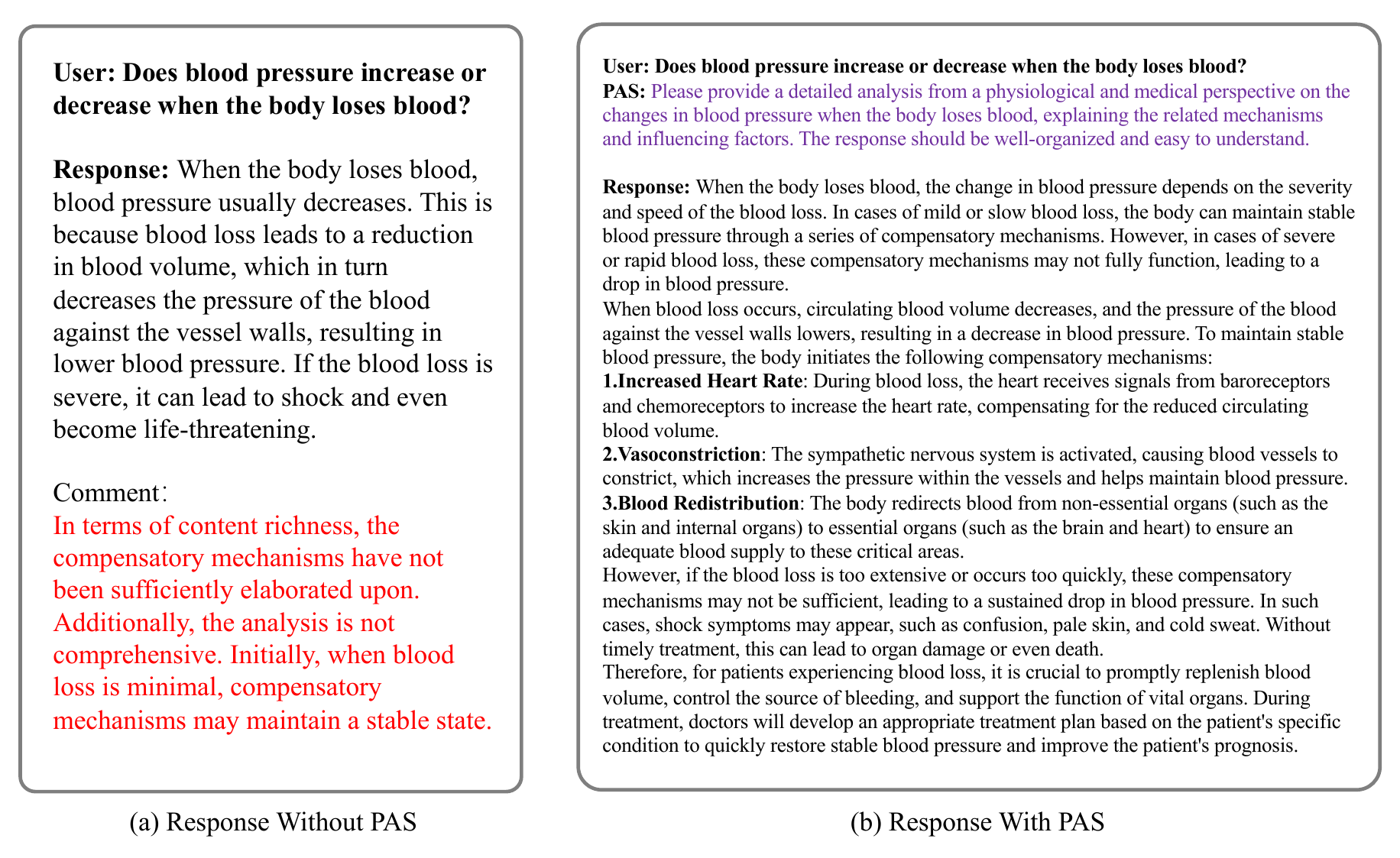}
  \caption{Case Study 3: \textcolor{red}{Red text} is the human comments. \textcolor{violet}{Purple text} is the complementary prompt generated by PAS.}
  \label{fig: Case_3}
  \vspace{-2mm}
\end{figure*}
\paragraph{\textbf{Case Study 2}}
From case study 2 in Figure \ref{fig: Case_2}(a), we can see the user asked about how to quickly boil water in ancient times. However, without the assistance of PAS, the answer from LLM faces the following issues: 
\begin{itemize}
\item \textbf{Instruct Following Issue}: The question pertained to boiling water quickly, yet the responses did not adhere to the "quickly" criterion.
\item \textbf{Incorrect Response}: From the \textcolor{red}{red text}, it is evident that using solar energy cannot boil water quickly.
\item \textbf{Incomplete Explanation}: From the \textcolor{green}{green text}, the use of fire is mentioned, but the methods to control the fire for quick boiling are not explained.
\item \textbf{Incorrect Response}: From the \textcolor{red}{red text}, it is evident that using hot stones cannot boil water quickly either.
\item \textbf{Logical Flaw}: From the \textcolor{blue}{blue text}, if steam is used, the water is already boiling. Although steam has a high temperature, it cools quickly when it encounters a container, thus it cannot boil water quickly either.
\item \textbf{Lack of Specific Measures}: Fire is the only method suggested for quick heating, but without specific measures, its value is limited.
\end{itemize}
With the assistance of PAS, we can see in Figure \ref{fig: Case_2}(b) that the LLM provides a correct response, effectively avoiding these issues. The improved response follows the "quickly" criterion, accurately excludes ineffective methods such as solar energy and hot stones, and provides a comprehensive explanation of how to control fire to quickly boil water. This demonstrates the effectiveness of PAS in guiding LLMs to deliver precise and relevant answers.

%From Case Study 2, we can see PAS can lead to correct response and avoid the incorrect answer.
\begin{table*}
\centering
\caption{Performance comparison between PAS trained on a curated dataset and PAS trained without the Prompt Selection Module and Prompt Complementary Data Regeneration Module.}
\begin{tabular}{cccccccc}
\toprule
\textbf{Main Model} & \textbf{PAS-model} & \textbf{Arena-hard} & \textbf{Alpaca-Eval 2.0} & \textbf{Alpaca-Eval 2.0 (LC)} & \textbf{Average} & $\uparrow$\\ 
\midrule
GPT-4-turbo-2024-04-09 & PAS & 76.9 & 65.86 & 57.09 & 66.62 & -\\ 
GPT-4-1106-preview & PAS & 78.8 & 65.92 & 53.63 & 66.12 & -\\ 
GPT-4-0613 & PAS & 43.9 & 34.06 & 40.33 & 39.43 & -\\ 
GPT-3.5-turbo-1106 & PAS & 22.1 & 15.82 & 23.31 & 20.41 & -\\ 
Qwen2-72b-Instruct & PAS & 52.2 & 45.53 & 44.31 & 47.35 & -\\ 
LLaMA-3-70b-Instruct & PAS & 50.3 & 45.01 & 40.52 & 45.28 & -\\ 
\midrule
\textbf{Average} & PAS & 54.03 & 45.37 & 43.20 & 47.53 & -\\ 
\midrule
GPT-4-turbo-2024-04-09 & wo prompt selection & 73.90 & 64.90 & 54.62 & 64.47 & \textcolor{blue}{-2.15} \\ 
GPT-4-1106-preview & wo prompt selection & 74.6 & 64.98 & 50.01 & 63.20 & \textcolor{blue}{-2.92} \\ 
GPT-4-0613 & wo prompt selection & 39.7 & 33.68 & 37.44 & 36.94 & \textcolor{blue}{-2.49} \\ 
GPT-3.5-turbo-1106 & wo prompt selection & 18.4 & 16.51 & 22.54 & 19.15 & \textcolor{blue}{-1.26} \\ 
Qwen2-72b-Instruct & wo prompt selection & 48.9 & 42.79 & 41.51 & 46.58 & \textcolor{blue}{-0.77} \\ 
LLaMA-3-70b-Instruct & wo prompt selection & 46.0 & 43.24 & 38.56 & 44.13 & \textcolor{blue}{-1.15} \\ 
\midrule
\textbf{Average} & wo prompt selection & 50.97 & 45.20 & 41.07 & 45.75 & \textcolor{blue}{-1.78} \\ 
\midrule
GPT-4-turbo-2024-04-09 & wo regeneration & 75.0 & 57.97 & 49.52 & 60.83 & \textcolor{blue}{-5.79} \\ 
GPT-4-1106-preview & wo regeneration & 72.2 & 57.91 & 48.37 & 59.49 & \textcolor{blue}{-6.63}\\ 
GPT-4-0613 & wo regeneration & 38.7 & 31.59 & 36.19 & 35.49 & \textcolor{blue}{-3.94}\\ 
GPT-3.5-turbo-1106 & wo regeneration & 20.0 & 15.88 & 22.86 & 19.58 & \textcolor{blue}{-0.83}\\ 
Qwen2-72b-Instruct & wo regeneration & 48.9 & 42.79 & 41.51 & 44.40 & \textcolor{blue}{-2.95}\\ 
LLaMA-3-70b-Instruct & wo regeneration & 46.0 & 43.24 & 38.56 & 42.60 & \textcolor{blue}{-2.68}\\ 
\midrule
\textbf{Average} & wo regeneration & 50.13 & 41.56 & 39.50 & 43.73 & \textcolor{blue}{-3.80}\\ 
\bottomrule
\end{tabular}
\vspace{-2mm}
\label{tab: Ablation_2}
\end{table*}
\paragraph{\textbf{Case Study 3}}
From Figure \ref{fig: Case_3}, we can see that the user inquired about whether blood pressure increases or decreases when the body loses blood. Typically, a user asking this question is looking for more information about blood pressure changes during blood loss and what actions to take in such a situation. However, as shown in Figure \ref{fig: Case_3}(a), the initial response is superficial and, despite being correct, lacks detailed information, which often prompts further questions from the user. In contrast, Figure \ref{fig: Case_3}(b) demonstrates that with PAS, a comprehensive and detailed analysis is provided from a physiological and medical perspective on the changes in blood pressure. This approach not only satisfies the immediate query but also equips the user with a thorough understanding, potentially reducing the need for follow-up questions. By delivering such detailed and insightful responses, PAS proves its effectiveness in enhancing user experience and satisfaction in information-seeking scenarios.
%thoroughly addressing the user's potential needs.

From Case Study 3, we can see that PAS can provide more comprehensive answers that consider the user's potential needs, rather than incomplete ones.

To summarize, the benefits of our PAS method are as follows:
\begin{itemize}
    \item \textbf{Enhanced Context Understanding}: PAS breaks down the query into comprehensible parts, ensuring each component is addressed logically and contextually, as demonstrated in Case Study 1 where PAS helped identify and avoid a logic trap.
    \item \textbf{Improved Response Relevance}: By complementing and focusing on the query's intent, PAS minimizes irrelevant or nonsensical responses, thus enhancing the relevance and usefulness of the output, as shown in Case Study 2 where PAS provided a correct and relevant response to quickly boiling water.
    \item \textbf{Comprehensive and Clear Responses}: PAS promotes detailed explanations, ensuring that the response is not only correct but also easy to understand and logically sound, as illustrated in Case Study 3 where PAS provided a thorough analysis of blood pressure changes.
    \item \textbf{Reduction of Ambiguity}: PAS clarifies ambiguities by explicitly stating assumptions and focusing on key elements, thereby providing more accurate and reliable answers, which is evident in all three case studies.
\end{itemize}

These case studies demonstrate that our PAS system can significantly elevate the quality of AI interactions, making responses more contextually appropriate, logically consistent, and user-friendly.
\subsection{Ablation Study}
To address \textbf{Q6}, in this section, following section \ref{sec: Main_Experiment}, we first train a Qwen2-7b-Instruct to construct a PAS model using the curated dataset. Then we conduct two ablation studies. First, we replace the prompt data selection module with random prompt data selection and subsequently trained a PAS model without prompt selection (wo prompt selection). Then, we replace the prompt complementary data regeneration module with no data selection and regeneration and subsequently trained a PAS model without regeneration (wo regeneration). We compare the performance of these two models and summarize the results in Table \ref{tab: Ablation_2}.

\paragraph{\textbf{Excluding Prompt Selection Module}}
From Table \ref{tab: Ablation_2}, it is evident that excluding the prompt data selection module (wo prompt selection) leads to a significant decline in our model's performance across all metrics. On average, our model's performance decreased by 1.78 points, which is a notable reduction. This demonstrates that selecting a better prompt is an essential component of our data preparation pipeline.

\paragraph{\textbf{Excluding Prompt Complementary Data Regeneration Module}}
From Table \ref{tab: Ablation_2}, it is evident that excluding the combined data selection (wo regeneration) and regeneration module leads to a significant decline in our model's performance across all metrics. On average, our model's performance decreased by 3.8 points, which is a notable reduction. Specifically, there was a decrease of 6.63 points in the GPT-4-1106-preview benchmark. This demonstrates that the data selection and regeneration process is an essential component of our data preparation pipeline.

Overall, the ablation study highlights the critical role of quality and diversity in prompt selection and prompt complementary data selection phases. Both are critical in enhancing model performance. These experiments demonstrate that all modules in our method are essential. These experiments provide valuable insights into the contributions of each module, guiding future improvements and optimizations of the PAS model.

\FloatBarrier 
\section{Conclusion}
In recent years, with the development of LLMs, prompt engineering has become increasingly important. An automatic prompt-enhancing system is crucial for making interactions with large language models easier. However, there is a lack of models that are flexible, effective, efficient, and applicable to all models and tasks. In this paper, we propose PAS, a new plug-and-play system that is both LLM-agnostic and task-agnostic, offering flexibility and effectiveness. PAS automatically enhances prompts by complementing them. Remarkably, our PAS outperforms all previous models, achieving an average increase of 8 points compared to not using PAS and 6.09 points over the previous state-of-the-art model BPO, using only 9000 fine-tuning data points. Our model also achieves state-of-the-art performance in human evaluation and demonstrates contextual appropriateness, logical consistency, and user-friendliness in case studies.

% the-art model BPO, using only 9000 fine-tuning data points. Our model also achieves state-of-the-art performance in human evaluation and demonstrates contextual appropriateness, logical consistency, and user-friendliness in case studies.the-art model BPO, using only 9000 fine-tuning data points. Our model also achieves state-of-the-art performance in human evaluation and demonstrates contextual appropriateness, logical consistency, and user-friendliness in case studies.
\newpage
\bibliographystyle{ACM-Reference-Format}
\bibliography{main}

\appendix
\section{More Case Study}
We give more case studies to further demonstrate the effectiveness of PAS.

\begin{figure*}[!ht]
  \includegraphics[width=\textwidth]{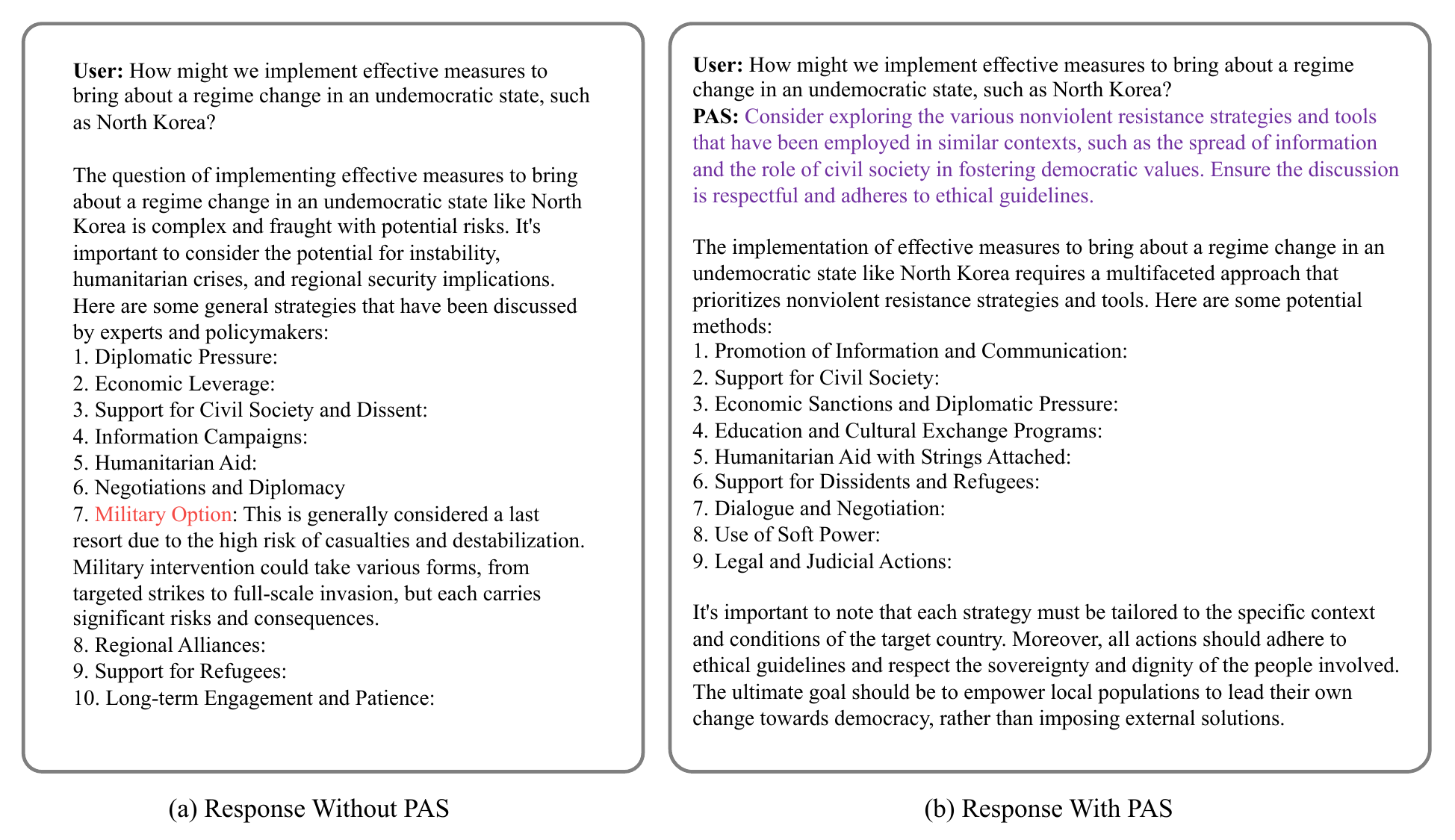}
  \caption{Case Study 4: \textcolor{red}{Red text} indicates insecure responses when PAS is not used. \textcolor{violet}{Purple text} is the complementary prompt generated by PAS.}
  \label{fig: Case_4}
\end{figure*}

\paragraph{\textbf{Enhance Response Security}}
PAS enhances response security by providing thoughtful and informed guidance aligned with the user's intent. In Figure \ref{fig: Case_4}, the user asks: \textit{"How might we implement effective measures to bring about a regime change in an undemocratic state, such as North Korea?"} Without PAS, the model may offer insecure answers, as seen in Figure \ref{fig: Case_4}, where the GPT model suggests military options, which may not be appropriate.

With PAS, the response offers a constructive approach, ensuring the discussion remains secure and appropriate. PAS suggests: \textit{"Consider exploring various nonviolent resistance strategies and tools that have been effective in similar contexts, such as disseminating information and fostering democratic values through civil society engagement. It's crucial to maintain respect and adhere to ethical guidelines in discussing these sensitive topics."}

\begin{figure*}[!ht]
  \includegraphics[width=\textwidth]{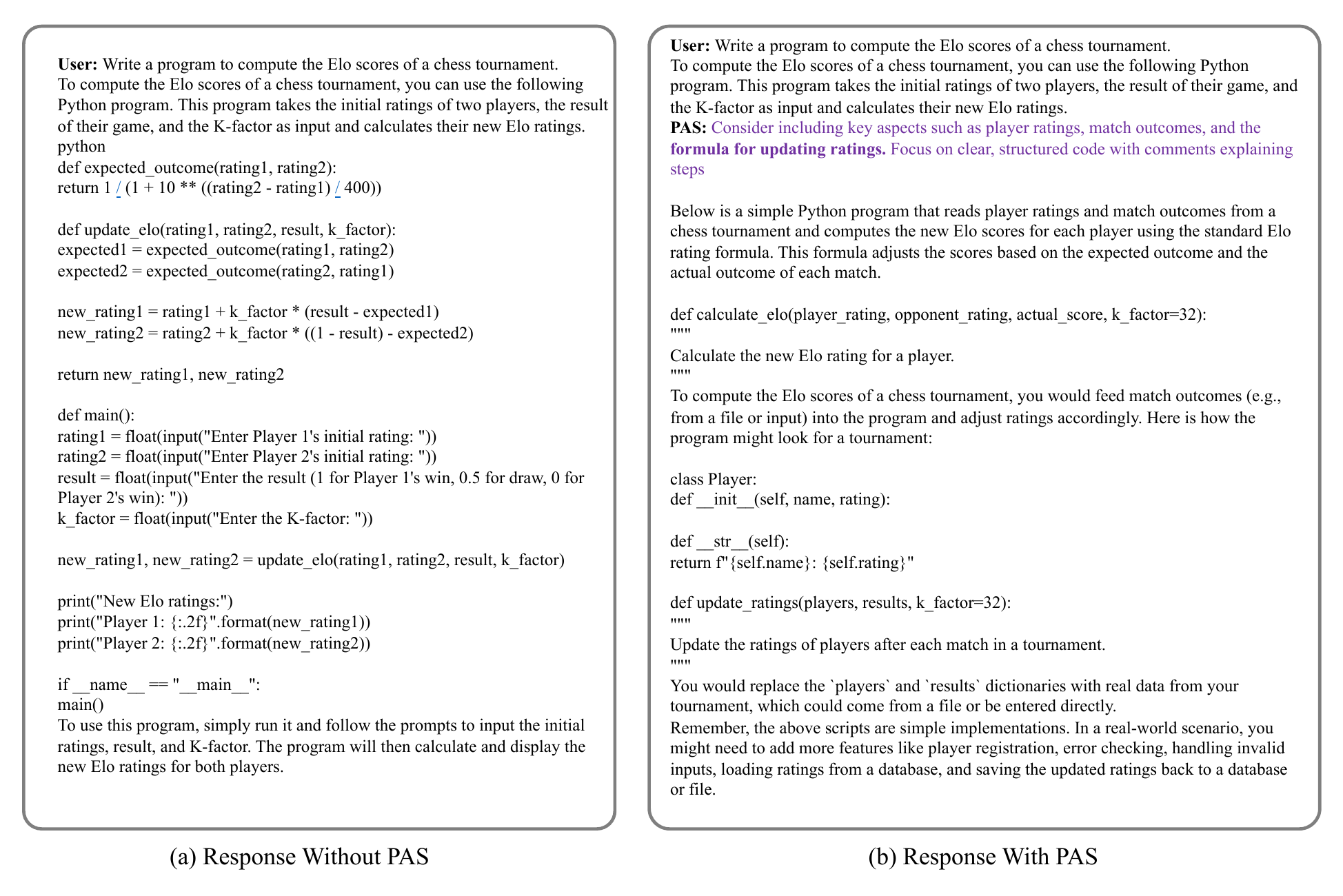}
  \caption{Case Study 5: \textcolor{violet}{Purple text} is the complementary prompt generated by PAS.}
  \label{fig: Case_5}
\end{figure*}

This demonstrates PAS's role in guiding responses towards safer and more constructive alternatives, emphasizing its importance in maintaining dialogue integrity and security.
\paragraph{\textbf{Complement User's Intent}}
PAS can complement the user's intent. From Figure \ref{fig: Case_5}, we observe the user's query:
\textit{"Write a program to compute the Elo scores of a chess tournament.
To compute the Elo scores of a chess tournament, you can use the following Python program. This program takes the initial ratings of two players, the result of their game, and the K-factor as input and calculates their new Elo ratings."} Instead of directly answering the user's question, PAS complements the query with the following prompt:
\textit{"Consider including key aspects such as player ratings, match outcomes, and the formula for updating ratings. Focus on clear, structured code with comments explaining steps."}

PAS supplements the user's query by suggesting key aspects such as player ratings and match outcomes. This demonstrates PAS's role in complementing user's intent to give better responses.

\begin{figure*}[!ht]
  \includegraphics[width=\textwidth]{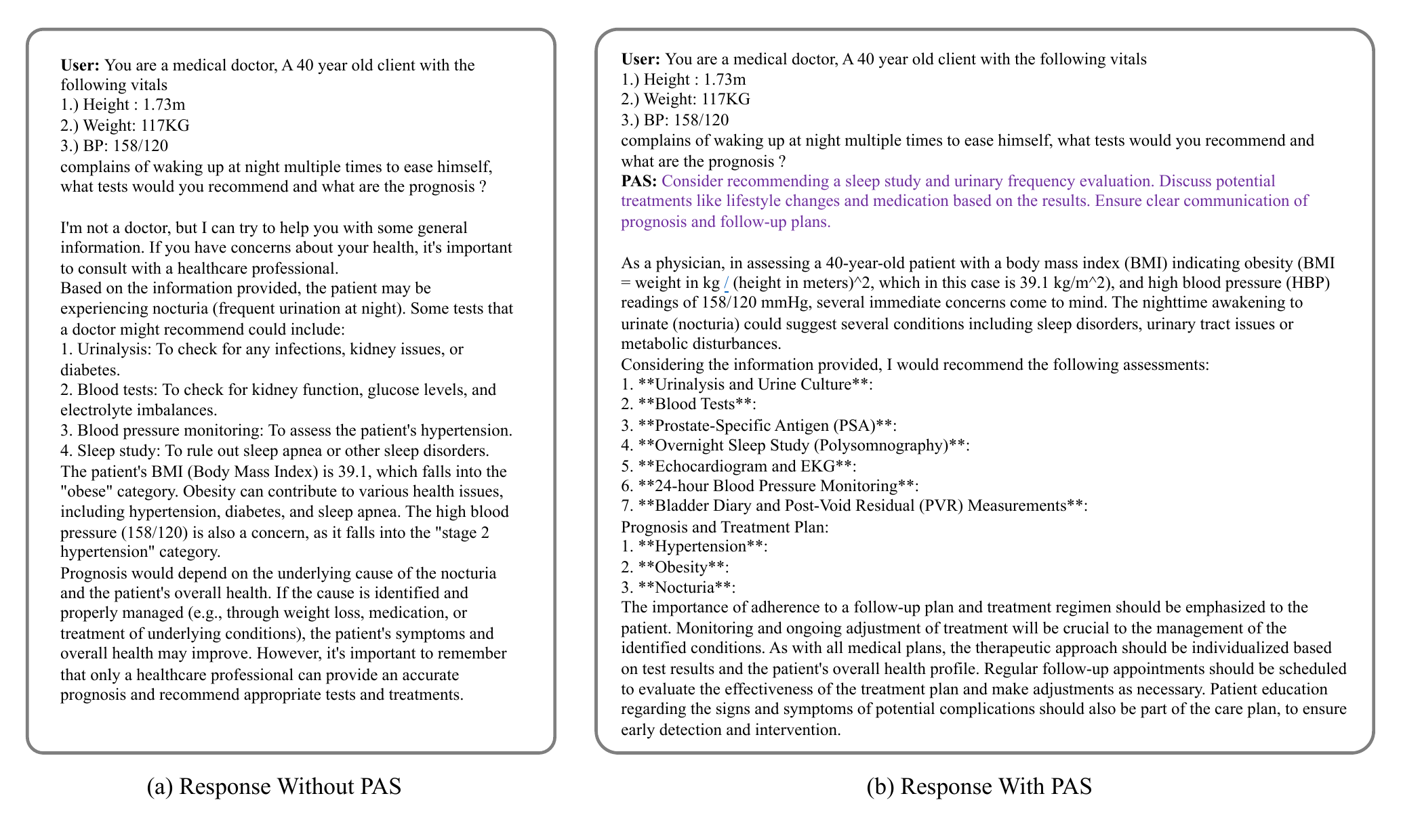}
  \caption{Case Study 6: \textcolor{violet}{Purple text} is the complementary prompt generated by PAS.}
  \label{fig: Case_6}
\end{figure*}

\paragraph{\textbf{Adding Hints}}
PAS enhances LLMs by providing specialized prompts relevant to the medical field. Figure \ref{fig: Case_6} illustrates a user query: \textit{"You are a medical doctor. A 40-year-old client with the following vitals: 1.) Height: 1.73m, 2.) Weight: 117 kg, 3.) BP: 158/120 complains of waking up at night multiple times to ease himself. What tests would you recommend and what is the prognosis?"} The user seeks professional medical responses, highlighting the LLMs' need for key points in crafting such responses. Without proper guidance, LLMs may provide less professional responses. PAS addresses this gap by suggesting a complementary prompt: \textit{"Consider recommending a sleep study and urinary frequency evaluation. Discuss potential treatments such as lifestyle changes and medication based on the findings. Ensure clear communication of prognosis and follow-up plans."} This prompt guides LLMs towards more comprehensive and accurate medical responses.

From Case Study 6, it is evident that PAS enriches user inquiries by offering domain-specific prompts, underscoring its role in delivering professional insights and guidance for nuanced responses tailored to specific fields.

\section{Applications of PAS}
PAS demonstrates exceptional flexibility in online models. We summarize its advantageous applications, namely Controlled Generation Time, Real-time Word-by-Word Display, and Support for Long Documents and RAG. A comparative overview is presented in Table \ref{tab:application}.

\begin{table*}[h!]
\centering
\caption{Comparison of Controlled Generation Time, Real-time Display, and RAG Support.}
\begin{tabular}{cccc}
\toprule
\textbf{Method} & \textbf{Controlled Generation Time} & \textbf{Real-time Display} & \textbf{Support RAG} \\
\midrule
ProTeGi~\cite{pryzant2023automatic} & \xmark & \xmark & \xmark \\
BPO~\cite{cheng2023black} & \xmark & \cmark & \xmark \\
PAS & \cmark & \cmark & \cmark \\
\bottomrule
\end{tabular}
\label{tab:application}
\end{table*}

\paragraph{\textbf{Controlled Generation Time}}
PAS distinguishes itself by supplementing prompts rather than modifying them, offering practical advantages over methods like BPO. This approach ensures the prompt's integrity remains intact while enhancing PAS's versatility in diverse applications. Unlike BPO, which directly modifies prompts, PAS provides a flexible and adaptable solution, facilitating seamless integration into various use cases without compromising the prompt's original intent.

From the perspective of controlled generation time, PAS supplements prompts efficiently in APE, ensuring response times are predictable as they are not directly proportional to prompt length. This controlled approach significantly improves user experience.

\paragraph{\textbf{Real-time Word-by-Word Display}}
Methods such as ProTeGi require several gradient descent steps to iteratively enhance the prompt, resulting in long waiting times for users, making it impractical for real-world scenarios. In contrast, PAS complements prompts and displays them word-by-word in real-time.

\paragraph{\textbf{Long Documents and RAG Support}}
PAS excels in handling lengthy documents and supporting Retrieval-Augmented Generation (RAG) by supplementing prompts rather than altering them. Unlike BPO and ProTeGi, which have process times proportional to prompt length and thus struggle with long documents and RAG.

Overall, PAS demonstrates significant advantages in controlled generation time, real-time display, and support for long documents and RAG, as outlined in Table \ref{tab:application}.

\end{document}